\documentclass[Afour,sageh,times]{sagej}

\usepackage{amsmath,amsfonts,amssymb}
\usepackage{bm}
\usepackage{graphicx}
\usepackage{booktabs}
\usepackage{multirow}
\usepackage{array}
\usepackage{rotating}
\usepackage{siunitx}
\usepackage{float}
\usepackage{textcomp}
\usepackage{verbatim}
\usepackage[inline]{enumitem}
\usepackage{tikz}
\usetikzlibrary{matrix,positioning,calc}

\usepackage{algorithm}
\usepackage{algorithmic}
\usepackage{tabularx}

\usepackage[caption=false,font=footnotesize]{subfig}

\usepackage[colorlinks,bookmarksopen,bookmarksnumbered,
            citecolor=red,urlcolor=red]{hyperref}

\newcommand{\bs}{\boldsymbol}

\begin{document}

\runninghead{Beber et al.}

\title{Autonomous Ergodic Robotic Palpation for Elastic Mapping of Curved Soft Tissues}

\author{Luca Beber\affilnum{1,2,3}*,
Cem Bilaloglu\affilnum{4,5}*,
Edoardo Lamon\affilnum{1,2},
Matteo Saveriano\affilnum{1,6},
Daniele Fontanelli\affilnum{1,6},
Luigi Palopoli\affilnum{1,2} and
Sylvain Calinon\affilnum{4,5}
}

\affiliation{
*Luca Beber and Cem Bilaloglu contributed equally to this work.
\affilnum{1}Interdepartmental Robotics Labs (IDRA), Universit\`a di Trento, 38123 Trento, Italy\\
\affilnum{2}Department of Information Engineering and Computer Science, Universit\`a di Trento, 38123 Trento, Italy\\
\affilnum{3}PhD Programme in Robotics and Intelligent Machines (DRIM), University of Genoa, 16126 Genoa, Italy\\
\affilnum{4}Idiap Research Institute, 1920 Martigny, Switzerland\\
\affilnum{5}\'Ecole Polytechnique F\'ed\'erale de Lausanne (EPFL), 1015 Lausanne, Switzerland\\
\affilnum{6}Department of Industrial Engineering, Universit\`a di Trento, 38123 Trento, Italy
}

\corrauth{
Luca Beber, Department of Information Engineering and Computer Science,
Universit\`a di Trento, 38123 Trento, Italy.
}

\email{luca.beber@unitn.it}

\begin{abstract}
  Autonomous robotic palpation offers repeatable, quantitative
  estimates of tissue properties for stiff inclusion
  localization. However, existing methods often rely on prior mesh
  models or planar surface assumptions that do not generalize to
  complex human anatomy. We propose an ergodic palpation framework that
  operates directly on curved soft surfaces using point cloud
  data. Central to our approach is the Laplace--Beltrami operator,
  which grounds both online estimation and exploration in the surface
  intrinsic geometry. We estimate local viscoelastic parameters using
  an Unscented Kalman Filter (UKF) and fuse them into a surface-aware
  Gaussian Process (GP). This probabilistic model drives a
  diffusion-based ergodic motion planner, generating a continuous
  exploration trajectory that feeds the new informative samples back
  to the GP in a closed loop. Throughout this process, a hybrid
  force-impedance controller regulates interaction forces to maintain
  contact. We evaluate the approach against existing methods in
  simulation and validate it in physical experiments on
  tissue-mimicking silicone phantoms with varying geometries and
  viscoelastic properties, demonstrating geometry-consistent mapping
  and stiff inclusion localization under contact-rich conditions.
\end{abstract}

\keywords{Robotic palpation, ergodic exploration, manifold-aware
Gaussian processes, elastic tissue mapping, medical robotics}

\maketitle

\section{Introduction}
\label{sec:introduction}
Palpation is a clinical procedure in which physicians assess possible anomalies from the mechanical response of the patient's tissues to touch. When performed manually, it is subjective and varies across practitioners, which limits diagnostic repeatability, particularly for small, deep-seated, or weakly contrasted lesions. Clinical workflows therefore complement palpation with imaging modalities such as mammography~\citep{Eisemann2025NationwideScreening} and ultrasound~\citep{Dan2024DiagnosticReview}, which are costly, require specialised staff and equipment, and can burden the patient (e.g., through ionising radiation).
Robotic palpation promises a more repeatable and objective interaction. Beyond revealing possible lesions, it can also quantify elasticity and viscosity~\citep{palacio2015quantitative}, which are useful for instance to regulate robot--tissue interaction forces in robot-assisted minimally invasive surgery and ultrasound-guided procedures.

\begin{figure}[ht!]
    \centering
    \includegraphics[width=0.95\linewidth]{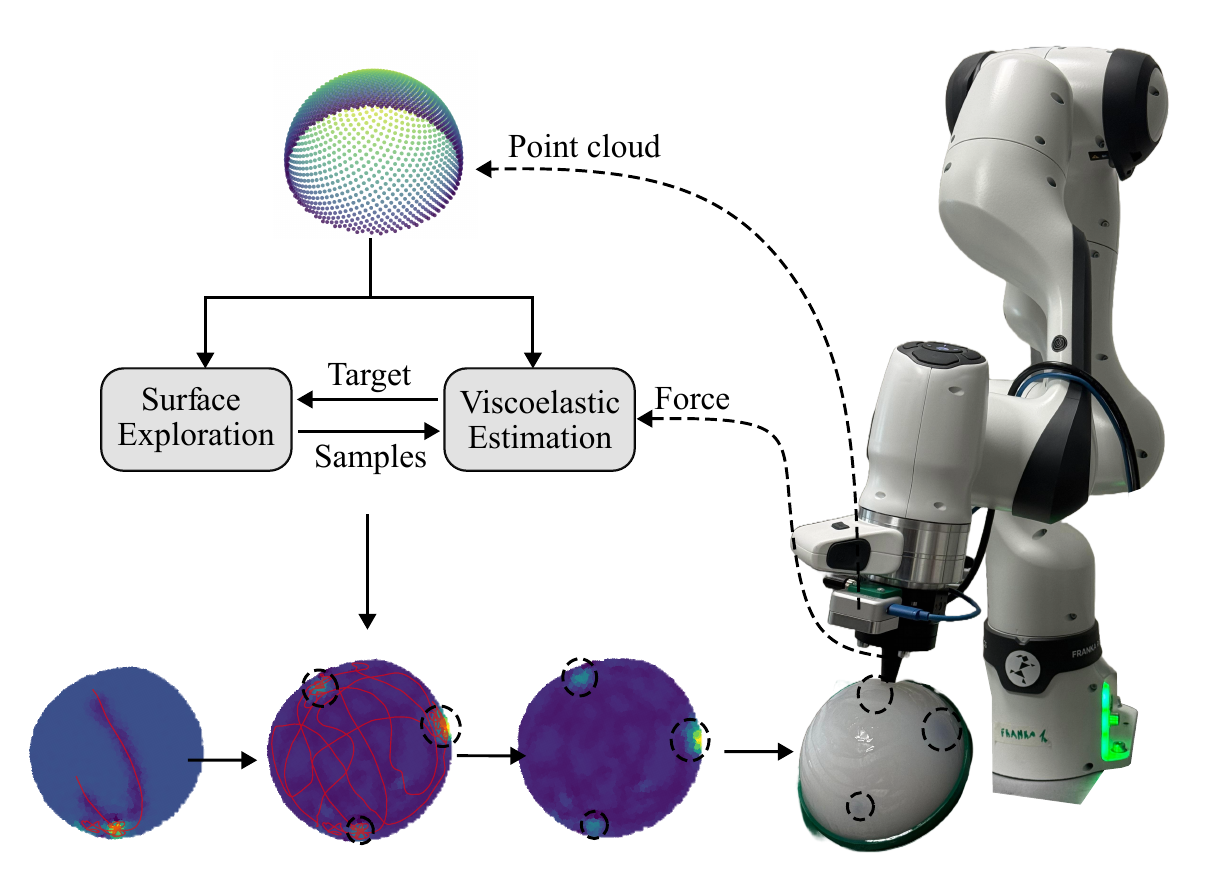}
    \caption{High-level overview of the framework. (\emph{top left}) The surface is sampled as a point cloud and processed to construct a surface-aware domain for exploration and viscoelastic estimation. Exploration collects samples for viscoelasticity estimation, while the planner returns target regions for future exploration. (\emph{bottom left}) Evolution of the estimated elastic map during palpation. The red curve shows the exploration trajectory, and the yellow regions inside dashed circles indicate inclusions detected by our method. (\emph{right}) Snapshot from the experiments of a robotic arm performing autonomous palpation on a hemisphere phantom.}
    \label{fig:overview}
\end{figure}

Approaches to tissue mechanical characterisation fall into imaging-based elastography and contact-based sensing. Ultrasound~\citep{Zhang2025InnovativeLesions} and magnetic resonance~\citep{Sack2023MagneticImaging} elastography estimate spatial stiffness distributions, but integrating them into robot-mediated, contact-rich procedures requires dedicated hardware, accurate registration between imaging and robot frames, and workflow adaptations.
Contact-based approaches instead infer tissue properties from force and deformation at the tool-tissue interface, providing local mechanical feedback in real time. Force feedback during probing has been shown to reduce applied forces and procedure time~\citep{Trejos2009Robot-assistedLocalization,Enayati2016HapticsBenefits}, and force-based estimation can recover meaningful stiffness information without specialised tactile arrays~\citep{Haddadi2012real,Roveda2022SensorlessEstimation}. This motivates using a single wrist-mounted force-torque (F/T) sensor both to regulate the interaction and to estimate local viscoelastic properties.

Palpation sampling strategies have evolved from exhaustive grid probing~\citep{McKinley2016AnResection, Zevallos2018AInformation} to information-driven exploration based on Bayesian Optimization (BO)~\citep{Garg2016TumorSampling,Ayvali2017Utility-GuidedAbnormalities,Salman2018Trajectory-OptimizedSurgery}, which reduces the number of probing actions needed to localize abnormalities, and on ergodic exploration~\citep{Ayvali2017Utility-GuidedAbnormalities,beber2026autonomous}, which generates continuous, dynamically feasible trajectories whose spatial statistics match a target information distribution. However, these methods are typically formulated and validated on planar or locally flat surfaces~\citep{Garg2016TumorSampling} and assume rectangular exploration patches~\citep{Nichols2015MethodsPalpation,Garg2016TumorSampling,Ayvali2017Utility-GuidedAbnormalities,beber2026autonomous}. This limits their applicability to curved anatomical surfaces and to irregular or non-convex admissible contact regions arising from safety constraints, anatomy, or kinematic limits.

We address these limitations with an autonomous palpation framework for elastic mapping on curved soft surfaces represented as point clouds, without requiring an anatomical mesh, a volumetric model, or a global surface parameterization.  \autoref{fig:overview} provides an overview of the framework. Local viscoelastic parameters are estimated online from force-displacement measurements using our previous force-based framework~\citep{Beber2025Force-BasedArm}, in which a contact model based on the Dimensionality Reduction Method (DRM)~\citep{Popov2015MethodFriction} is embedded in a Kalman filter. These estimates are fused into a manifold-aware Gaussian Process (GP)~\citep{Borovitskiy2020MaternManifolds}
defined directly over the point cloud, whose posterior drives an ergodic planner in the spirit of HEDAC~\citep{Bilaloglu2025TactileSurfaces}. The ergodic planner, point cloud HEDAC (PC-HEDAC), is updated at a high rate and requires neither surface flattening nor discrete waypoint planning. Its Expected Information Density (EID) combines attraction towards regions of high estimated apparent stiffness with exploration of regions of high posterior uncertainty, progressively refining the map around mechanically informative areas, according to the prescribed ergodic metric.
The framework is evaluated through systematic simulation studies and repeated physical experiments.
The results demonstrated that the proposed method achieves lower reconstruction RMSE than all evaluated baselines under the same travel budget and requires approximately five times less travel than BO to reach the same RMSE. In addition, the proposed stopping criterion reduces exploration time by approximately $30\%$ compared with full exploration runs, with comparable reconstruction and localization performance.

 To summarize, the main contributions of this work are:

\begin{itemize}[nosep]
    \item A closed-loop robotic palpation framework coupling force-based
    viscoelastic estimation, intrinsic probabilistic mapping, and
    continuous exploration directly on non-planar point cloud surfaces.

    \item A shared intrinsic geometric representation for mapping and
    exploration, using both the spectral representation of the Laplace--Beltrami operator to construct the
    heat-kernel GP and the full discrete representation for the diffusion-based
    planner, without global surface parameterization or spectral
    truncation of the planner.

    \item A systematic evaluation on simulated and physical curved
    surfaces, including intrinsic versus Euclidean mapping, comparisons with alternative exploration strategies, physical reference mapping, localization of stiff regions, and a stopping criterion based on the defined ergodic metric.
\end{itemize}

\begin{table*}[ht!]
\caption{Comparison of robotic palpation methods by mapping output,
exploration, contact control, and surface representation.}
\label{tab:palpation_mapping_sota_rotated}
\centering
\small
\setlength{\tabcolsep}{4pt}
\renewcommand{\arraystretch}{1.05}
\begin{tabularx}{\linewidth}{
    @{}
    >{\raggedright\arraybackslash}p{0.10\linewidth}
    >{\raggedright\arraybackslash}X
    >{\raggedright\arraybackslash}X
    >{\raggedright\arraybackslash}X
    >{\raggedright\arraybackslash}X
    @{}
}
\toprule
\textbf{Method} &
\textbf{Mapping / output} &
\textbf{Exploration} &
\textbf{Contact control / execution} &
\textbf{Surface representation} \\
\midrule

\citet{Nichols2015MethodsPalpation} &
Feature-based inclusion segmentation &
Grid sampling with local refinement &
Position control; force threshold &
Bounded surface patches \\

\citet{Garg2016TumorSampling} &
GP stiffness map and tumor boundary &
Uncertainty-driven probing &
Position control &
2D parameterization; Euclidean covariance \\

\citet{Ayvali2017Utility-GuidedAbnormalities} &
GP utility and stiffness fields &
Utility-guided BO / SMC &
Hybrid force--motion control &
Parameterized domain \\

\citet{Yan2021FastPalpation} &
GP stiffness map; tumor boundary and depth &
BO probing and tactile sliding &
Pressing and sliding &
Planar or weakly curved regions \\

\citet{Uppuluri2024SeeBelow:Exploration} &
Vision-assisted tumor reconstruction &
BO probing and contour following &
Hybrid force--impedance control &
Vision-derived mesh \\

\citet{Yue2025BimodalSegmentation} &
GP stiffness inference; OCT boundaries &
Bayesian sampling and tactile sliding &
Rotational and sliding palpation &
Planar samples \\

\citet{beber2026autonomous} &
GP elasticity map and lesion segmentation &
Continuous ergodic exploration &
Cartesian position control &
Planar rectangular domain \\

\midrule
\textbf{Ours} &
Intrinsic GP elasticity map &
Continuous ergodic exploration on point clouds &
Hybrid force--impedance control &
Curved point clouds; no global parameterization \\
\bottomrule
\end{tabularx}
\end{table*}

\section{Related Works}
\label{sec:related_works}

\autoref{tab:palpation_mapping_sota_rotated} summarizes the main approaches discussed below.

\subsection{Bayesian Optimization and GP-Based Palpation}
Early autonomous palpation systems relied on grid-based~\citep{6094786} or raster-like probing~\citep{McKinley2016AnResection,Zevallos2018AInformation}. These strategies provide predictable coverage and remain effective when a dense map of the whole surface is required, but they allocate measurements uniformly, regardless of the observed response or uncertainty, and can spend substantial time probing homogeneous tissue.
BO and related GP-based active sensing strategies were therefore introduced to reduce the number of probing actions required for tumor localization and stiffness mapping. They formulate exploration as a sequence of discrete sampling decisions driven by an acquisition function, typically without accounting for the robot's continuous-time dynamics or contact constraints. \citet{Garg2016TumorSampling} showed that acquisition design affects both peak detection and reconstruction quality, with Upper Confidence Bound (UCB) favoring regions of high predicted stiffness while accounting for posterior uncertainty.
\citet{Ayvali2017Utility-GuidedAbnormalities} used Expected Improvement (EI) to jointly address registration and elasticity mapping with priors and motion penalties.
For boundary reconstruction, GP-based level-set estimation (LSE) selects measurements to identify where an unknown function exceeds a threshold~\citep{Gotovos2013Active}, while \citet{Nichols2015MethodsPalpation} segmented stiff inclusions with probabilistic models such as Markov random fields. \citet{Yan2021FastPalpation} observed that BO localizes stiff regions efficiently but requires additional mechanisms to delineate boundaries, and coupled localization with a subsequent sliding-based boundary tracing phase. Such staged pipelines, however, depend critically on the reliability of intermediate estimates (e.g., centroid localization).

Other approaches perform continuous force-controlled palpation while estimating tissue properties along the executed trajectory.
\citet{Chalasani2018PreliminaryStiffness} used GPs to jointly estimate online tissue geometry and stiffness,
although the autonomous experiments followed a preplanned trajectory rather than adapting to the evolving stiffness map.
Trajectory-optimized sensing actively plans palpation paths from the current probabilistic estimate~\citep{Salman2018Trajectory-OptimizedSurgery}, and online replanning has been used to map the 3D geometry of subsurface anatomy~\citep{Cho2024EfficientSurgery}. These methods, however, rely on repeated trajectory optimization or graph search over a finite horizon, whose computational burden grows as the map is updated online. Further systems address 3D reconstruction and subsurface profiling, e.g., SeeBelow for tumor reconstruction and contour following~\citep{Uppuluri2024SeeBelow:Exploration} and bimodal tactile tomography for microstructure profiling~\citep{Yue2025BimodalSegmentation}.

\subsection{Tactile Ergodic Exploration}
Ergodic control generates trajectories whose visitation statistics match a prescribed information distribution while explicitly accounting for system dynamics~\citep{Mathew2011MetricsSystems,Miller2016ErgodicInformation}. The most widely adopted formulation, Spectral Multiscale Coverage (SMC), uses a Fourier-based metric that prioritizes coarse coverage before refining finer details. Ergodic trajectory optimization achieves a lower final ergodic metric over a finite horizon but is computationally demanding, limiting real-time replanning~\citep{Miller2013TrajectoryExploration}.

Ergodic control is particularly suited to contact-based tasks, where measurements are collected along the executed motion and the feasibility of transitions between sensing locations cannot be neglected. \citet{Abraham2017ErgodicEstimation} combined SMC with binary contact sensing for shape estimation, showing that information-optimal motion can compensate for limited sensing bandwidth. Later work accounted for sensor or tool footprints~\citep{Bilaloglu2025TactileSurfaces,kwonVolumetricErgodicControl2025} and the whole surface of a manipulator~\citep{Bilaloglu2023Whole-BodyDiffusion}. In palpation, \citet{Ayvali2017Utility-GuidedAbnormalities} reported inferior localization for SMC compared with BO under strong priors, indicating that both the ergodic formulation and real-time updates are crucial. Our previous work~\citep{beber2026autonomous} combined HEDAC-based exploration with GP regression on planar, rectangular domains, a Euclidean formulation that does not naturally extend to curved surfaces with irregular boundaries.

Extending spectral ergodic control beyond planar domains requires either a spectral basis on the underlying geometry or a parameterization of the surface. Laplacian eigenfunctions, which generalize Fourier series to Riemannian manifolds, have enabled Fourier-based metrics on spheres and tori~\citep{jacobsMultiscaleSurveillanceRiemannian2010}, planar domains with complex boundaries~\citep{xuMeasurePreservingFlows2024}, and watertight meshes~\citep{dongErgodicExplorationMeshable2025}. Alternatively, \citet{Schneyer2026AnPreferences} parameterize the surface onto a planar domain, apply SMC there, and map the trajectory back, but such parameterizations introduce distortion that grows with surface and boundary complexity. HEDAC~\citep{Ivic2017Ergodicity-BasedField} instead uses the stationary heat equation to generate smooth guidance fields and can handle arbitrary boundaries~\citep{ivicConstrainedMultiagentErgodic2022} and surfaces~\citep{ivicMultiUAVTrajectoryPlanning2023}, but was initially limited to watertight meshes and offline planning due to its computational cost.

In realistic palpation, watertight meshes of human tissue are typically unavailable without additional imaging, segmentation, and registration; the accessible surface is instead sensed intraoperatively, e.g., as a point cloud from an RGB-D camera. Recent ergodic planners use point clouds to represent the target distribution through maximum mean discrepancy~\citep{hughesErgodicTrajectoryOptimization} or flow matching with the Sinkhorn divergence~\citep{sunFlowMatchingErgodic2025}. However, the samples only define the target measure and do not encode the connectivity, geodesic structure, or boundaries of the contact surface. Without an explicit surface constraint, the robot may penetrate the tissue or drift away from it, causing excessive indentation or loss of contact. In contrast, our approach constructs an intrinsic surface operator directly from the point cloud and couples surface-constrained exploration with continuous force regulation.

GPs based on Euclidean distance are ill-suited to curved anatomy, since points close in 3D space may be far apart along the tissue surface. Manifold-aware kernels define spatial correlations intrinsically over the surface~\citep{Borovitskiy2020MaternManifolds} and support uncertainty-driven sampling on curved domains~\citep{Gao2019}, motivating their use for geometry-consistent stiffness mapping. However, such approaches often rely on additional sensing modalities or complex forward models beyond force-only interaction.

\subsection{Positioning of This Work}
The closest work to ours is the tactile ergodic controller based on HEDAC  of~\citet{Bilaloglu2025TactileSurfaces}, developed for surface cleaning. While both methods work on point cloud inputs, the main differences are represented by the domain representation (i.e. the target distribution) and by the planner goal. There, exploration is driven by a coverage error derived from vision and acquired once before execution, so the target distribution is known in advance and remains fixed. In our method for palpation, by contrast, no prior information on the mechanical properties is available: the informative regions are unknown and can only be revealed through contact. In this setting, the controller of~\citet{Bilaloglu2025TactileSurfaces} would reduce to uniform coverage of the surface. We instead close the perception--action loop by coupling force-based viscoelastic estimation with an intrinsic GP, whose posterior continuously updates the target distribution driving exploration. The robot thus builds a representation of the informative regions while palpating, using only an off-the-shelf F/T sensor.
This shifts the method from coverage-based exploration to integrated mechanical estimation and information-driven palpation on curved soft surfaces.

\section{Background}

This section presents the tools, algorithms and approaches used for
the solution development, namely the filtering approaches, the
Gaussian process modeling and the ergodic control basics.

\subsection{Continuous Estimation of Viscoelastic Properties}
\label{sec:ukf}
To enable spatially distributed characterization of soft materials,
local viscoelastic properties must be estimated continuously as the
probe moves across the surface. This requires an online estimator able
to update the contact state and the local elastic and viscous
parameters during motion, rather than relying on classical point-wise
indentation procedures performed at isolated quasi-stationary
locations. We rely on our previous force-based viscoelastic
characterization
framework~\citep{Beber2024TowardsArm,Beber2025Force-BasedArm}, where
the palpation model, excitation strategy, and online estimation
procedure were studied for robotic tissue-like material
characterization. In this work, this estimator provides the local
sensing module of the spatial exploration.  Using the Dimensionality
Reduction Method (DRM)~\citep{Popov2015MethodFriction}, and assuming a
spherical indenter, the total normal force $F_{z}$ can be written as:
\begin{equation}
\begin{aligned}
&F_{z}(d,\dot d) = \kappa d^{\frac{3}{2}}+\lambda d^{\frac{1}{2}}\dot d, \\
&\kappa =\frac{4}{3}\frac{E}{1-\nu^2}\sqrt{R}, \quad \lambda=\frac{4}{1-\nu}\eta\sqrt{R},
\end{aligned}
\label{eq:DRcompact_continuous}
\end{equation}
where $d$ and $\dot d$ denote penetration and penetration rate, $E$
the elasticity modulus, $\eta$ the viscosity coefficient, $\nu$ the
Poisson ratio, and $R$ the indenter radius. $\kappa$ and $ \lambda$ are used to define the relative elasticity and relative viscosity, respectively, that simplify the following notation.

Unlike static indentation approaches, the dynamic formulation allows treating penetration as an internal state. Therefore, the estimator does not require the explicit knowledge of the surface position and does not depend on potentially uncertain geometric calibration or force–threshold contact detection. Instead, $d$ is estimated online from the dynamic interaction, together with the viscoelastic parameters, ensuring full consistency of the estimation process.

\subsubsection{Expanded-State Dynamic Model}

To continuously track spatially varying viscoelastic properties, the estimator must accommodate time-varying elasticity and viscosity while guaranteeing sufficient responsiveness to local tissue changes. The state $\bs{x} =
[d, \kappa, \lambda, \dot d, \dot{\kappa}, \dot{\lambda}, \ddot{\kappa}, \ddot{\lambda}
]^{\top}$ incorporates penetration, its dynamics, and the viscoelastic parameters with their temporal derivatives.
Assuming an equivalent mass $m_I$, given by the weight of the indenter and the plate in front of the load cell acting along the contact normal, the penetration dynamics $\ddot d = \frac{1}{m_I}
(u - \kappa d^{\frac{3}{2}} - \lambda d^{\frac{1}{2}}\dot d)$, where $u$ is the measured force, follows from Newton’s law and the contact force~\eqref{eq:DRcompact_continuous}.

\subsubsection{Adaptive Fading Estimation}

The discretized model is embedded into an Unscented Kalman Filter (UKF). The robot end-effector velocity along the normal direction is used as a measurement, while the interaction force acts as the system input. Since the viscoelastic properties may change abruptly (e.g., when crossing a stiffer inclusion), a fading-memory strategy is introduced to improve responsiveness. Let $y_t$ denote the innovation term of the UKF. The covariance matrix is modified as $P_{t|t-1} \leftarrow \zeta P_{t|t-1}$,
where the fading factor $\zeta$ is updated online according to
\begin{equation}
\zeta =
\begin{cases}
\min(\zeta_{\max},\,\zeta+\Delta\zeta), & \|y_t\|>\theta, \\[4pt]
\max(1,\,\zeta-\Delta\zeta), & \text{otherwise}.
\end{cases}
\label{eq:fading_update}
\end{equation}
When the innovation magnitude exceeds the threshold $\theta$, the
filter temporarily increases its forgetfulness and prioritizes new
information. Conversely, when the system evolves in regions of
homogeneous tissue, $\zeta$ gradually returns to unity, restoring
nominal filtering behavior. The parameters $\varsigma_{\max}$, $\Delta\varsigma$, and $\theta$ were tuned in~\citet{Beber2025Force-BasedArm} to balance the ability to track changes in viscoelastic modulus with the noise induced by increasing $P_{t|t-1}$.

At each time step, the estimate $\hat{\kappa}(x,y)$ associated with the current exploration point is converted into spatially continuous elastic fields ($E$) by inverting~\eqref{eq:DRcompact_continuous},
resulting in an online map of the elastic and viscous properties over the palpated region.

\subsection{Gaussian Processes}
\label{sec:gp}

A Gaussian Process (GP) is a Bayesian non-parametric model for an unknown function $f(\bs{x})$, providing predictions together with associated uncertainty~\citep{Seeger2004GaussianLearning.,rasmussen2006gaussian}:
\begin{equation}
f(\bs{x}) \sim \mathcal{GP}\left(m(\bs{x}), k(\bs{x},\bs{x}')\right),
\end{equation}
where $m(\bs{x})$ and $k(\bs{x},\bs{x}')$ denote the mean and covariance functions, respectively.

Given observations $\bs{y}=[y_1,\dots,y_n]^\top$ at inputs
$\bs{X}=[\bs{x}_1,\dots,\bs{x}_n]^\top$, we assume additive Gaussian
measurement uncertainty:
\begin{equation}
\bs{y} = f(\bs{X}) + \bs{\epsilon}, \qquad
\bs{\epsilon} \sim \mathcal{N}\left(\bs{0},\sigma_n^2\bs{I}\right).
\end{equation}
Conditioning the GP on the observations yields a Gaussian posterior at test inputs $\bs{X}_*$:
\begin{equation}
p(f_* \mid \bs{X},\bs{y},\bs{X}_*)
= \mathcal{N}\left(\bs{\mu}_*,\bs{\Sigma}_*\right),
\end{equation}
where $f_*=f(\bs{X}_*)$, and the posterior mean and covariance are
\begin{align}
\bs{\mu}_* &= \bs{m}_* + \bs{K}_{*X}
\left(\bs{K}_{XX}+\sigma_n^2\bs{I}\right)^{-1}
\left(\bs{y}-\bs{m}_X\right), \label{eq:gp_mean} \\
\bs{\Sigma}_* &= \bs{K}_{**} - \bs{K}_{*X}
\left(\bs{K}_{XX}+\sigma_n^2\bs{I}\right)^{-1}
\bs{K}_{X*}. \label{eq:gp_cov}
\end{align}
Here, $\bs{m}_X=m(\bs{X})$, $\bs{m}_*=m(\bs{X}_*)$, and
$\bs{K}_{AB}=\bs{K}(\bs{A},\bs{B})$ denotes the covariance matrix evaluated between the input sets $\bs{A}$ and $\bs{B}$.

\subsection{Ergodic Control}
\label{sec:ergodic_control}
Consider a target spatial distribution $p(\bs{x})$ defined over a domain $\Omega$ that we wish to explore, and let $\bs{x}(t)$ denote the trajectory of a dynamical system~\citep{Ivic2017Ergodicity-BasedField,Miller2013TrajectoryExploration}. We define the coverage distribution at time $t$ as
\[
c(\bs{x}, t)
=
\int_{0}^{t}
\varphi\!\big(\bs{x}-\bs{x}(\tau)\big)\, d\tau,
\]
where $\varphi(\bs{r})$ denotes the sensor footprint centered at the system state. A Dirac distribution or Gaussian kernel are common choices for $\varphi(\bs{r})$.
The normalized time-averaged coverage is then given by
\[
\bar{c}(\boldsymbol{x}, t)=\frac{c(\boldsymbol{x}, t)}{\int_{\Omega} c(\boldsymbol{x}, t) d \boldsymbol{x}}.
\]
The dynamical system is said to be \emph{ergodic} with respect to the target distribution $p(\bs{x})$ if the time-averaged coverage converges to $p(\bs{x})$ as time goes to infinity, i.e.,
$
\lim_{t \to \infty} \bar c(\bs{x}, t) = p(\bs{x}).
$

\begin{figure*}[t]
    \centering
    \includegraphics[width=0.95\linewidth]{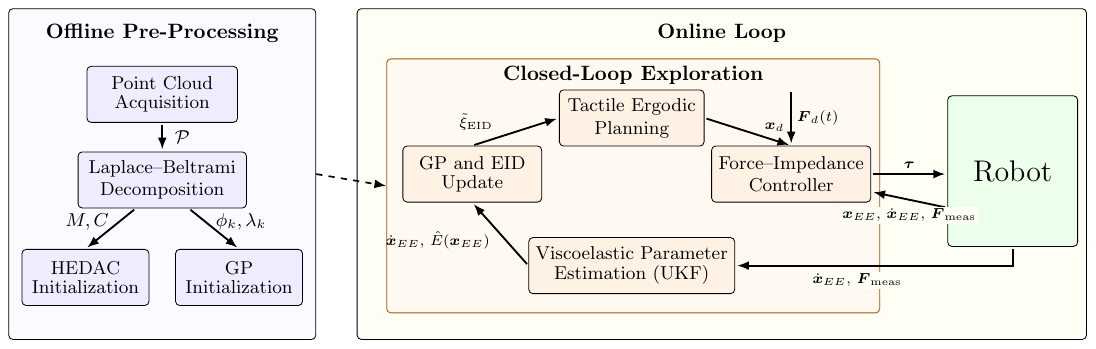}
    \caption{Detailed architecture of the proposed framework.
    An offline preprocessing stage acquires a point cloud representation of the workspace and performs a Laplace-Beltrami decomposition to initialize both the Gaussian Process (GP) model and the ergodic exploration basis.
    During online operation, a tactile ergodic control policy generates desired end-effector trajectories and interaction forces based on the current ergodic information density (EID).
    A Cartesian force-impedance controller ensures stable and compliant physical interaction with the environment, while an Unscented Kalman Filter (UKF) estimates local viscoelastic parameters from measured end-effector states and contact forces.
    The GP model and the EID are continuously updated, closing the perception-action loop and enabling adaptive, information-driven exploration.}
    \label{fig:framework}
\end{figure*}

\section{Methods}
\label{sec:methods}
An autonomous palpation framework on curved soft-tissue surfaces aims to reconstruct spatial elasticity distribution and localize stiff inclusions. The proposed framework consists of an offline geometric initialization phase followed by an online closed-loop exploration phase. The two phases share the Laplace--Beltrami formulation described in~\autoref{sec:surface_geometry}, which supports both GP covariance construction and diffusion-based ergodic exploration. The individual components are introduced here at the system level and described in detail in the subsequent subsections.
The overall architecture of the proposed framework is illustrated in~\autoref{fig:framework} and described as follows.

\paragraph{Offline Pre-Processing: Geometric Initialization}
Prior to online execution, the reference point cloud is used to construct a discrete approximation of the intrinsic surface geometry. From this representation, the weak Laplacian and mass matrices associated with the Laplace--Beltrami operator are computed together with the corresponding spectral decomposition.
The Laplace--Beltrami eigenpairs are used to construct the manifold-aware covariance kernel of the Gaussian process (GP), while the matrices are retained for online ergodic exploration.

\paragraph{Online Loop: Closed-Loop Exploration}
During online operation, four functional modules (explained in detail in \autoref{subsec:exploration_loop}) interact continuously to form a closed perception--action loop. Palpation measurements update the local material estimate, which is incorporated into the GP representation of the elasticity field. The GP posterior is then converted into an expected information density (EID), which defines the spatial objective of the ergodic exploration policy and consequently adapts the robot motion.

\subsection{Closed-Loop Exploration}
\label{subsec:exploration_loop}

\paragraph{Viscoelastic State Estimation}
As the robot interacts with the surface, interaction forces and end-effector motion along the surface normal are measured in real time. These signals are processed by an Unscented Kalman Filter, which jointly estimates the penetration state and the local viscoelastic parameters of the material, as detailed in \autoref{sec:ukf}. The local elasticity estimate and its corresponding surface location are supplied to the probabilistic mapping module, whereas viscosity is retained as an auxiliary estimate accounting for rate-dependent tissue response.

\paragraph{Probabilistic Mapping and Information Encoding}
The local elasticity estimates are incorporated into the GP to obtain a probabilistic representation of the spatial elasticity distribution over the surface. As described in \autoref{sec:surface_gp}, the GP covariance is defined intrinsically through the Laplace--Beltrami spectrum of the point cloud. The posterior mean and uncertainty are combined into the EID. Its exploitation component emphasizes positive elasticity deviations from the current spatial mean, thereby highlighting candidate stiffness peaks, whereas its exploration component favors regions with high posterior uncertainty.

\paragraph{Tactile Ergodic Planning}
The current EID defines the target spatial distribution of the ergodic
exploration policy. As detailed in \autoref{sec:ergodic_pc}, this
target is compared with the robot accumulated visitation to construct
an exploration deficit, which is diffused over the surface through the
Laplace--Beltrami operator. The gradient of the resulting potential
field directs the exploration agent toward regions that remain
under-visited relative to their current information density. The
resulting trajectory therefore balances repeated interaction with
regions of high estimated elasticity against exploration of uncertain
regions. The target end-effector position and corresponding surface
normal are then supplied as references for interaction control.

\paragraph{Interaction Control and Execution}
The motion references generated by the exploration policy are executed
through a Cartesian hybrid force--impedance controller
(in~\autoref{app:hybrid_fi_control}), which regulates the interaction
force while maintaining compliant contact with the surface. Tangential
motion is determined by the exploration policy, whereas the desired
end-effector (EE) interaction force along the surface normal is
prescribed as
\begin{equation}
    \bs{F}_d^{\mathcal{EE}}
    =
    \left[
        0,\,
        0,\,
        F_{\mathrm{bias}} + F_{\mathrm{amp}}\cos(\omega t),\,
        0,\,
        0,\,
        0
    \right]^\top,
    \label{eq:desired_force_exploration}
\end{equation}
where $\bs{F}_d^{\mathcal{EE}}$ is the desired force of the hybrid force-impedance controller expressed in the end-effector frame, $_{\mathrm{bias}}$ is the nominal contact force,
$F_{\mathrm{amp}}$ is the amplitude of the force modulation, and
$\omega$ is the excitation frequency. Force regulation along the surface normal serves two
purposes. First, it accommodates small discrepancies between the point
cloud representation and the physical surface. Second, it maintains a
consistent contact condition throughout the exploration, independently
from the local surface variations.

\subsection{Shared Intrinsic Geometric Representation}
\label{sec:surface_geometry}
Before exploration, a depth sensor acquires a reference point cloud
\begin{equation}
    \mathcal{P} := \{\bs{p}_i \in \mathbb{R}^3 \mid i = 1,\ldots,n_{\mathcal{P}}\},
\end{equation}
where $\bs{p}_i$ denotes the position of the $i$th surface sample. The point cloud provides the common geometric domain on which elasticity mapping and exploration are formulated.
The reference point cloud $\mathcal{P}$ is regarded as a discrete sampling of an underlying two-dimensional Riemannian manifold $\mathcal{M}$ embedded in $\mathbb{R}^3$. Although its points are represented by three-dimensional coordinates, the surface is intrinsically two-dimensional: locally, motion is restricted to directions tangent to the surface. The tangent plane therefore defines the directions of probe motion along the surface, whereas the surface normal defines the direction of indentation and force regulation. During palpation, probe-induced deformation is assumed to be predominantly normal and sufficiently localized to cause negligible changes in distances along the surface, or equivalently negligible in-surface strain. Under this approximation, the intrinsic geometry computed from the reference point cloud can be retained throughout exploration.

The Laplace–Beltrami operator \(\Delta_{\mathcal M}\) generalizes the Euclidean Laplacian to curved surfaces and provides the common geometric foundation for GP covariance construction and ergodic exploration. The diffusion of a scalar field $u$ over $\mathcal{M}$, starting from an initial distribution $u_0$, is governed by the heat equation:
\begin{equation}
    \frac{\partial u(\bs{x},s)}{\partial s}
    =
    \Delta_{\mathcal{M}}u(\bs{x},s),
    \qquad
    u(\bs{x},0)=u_0(\bs{x}),
    \label{eq:diffusion_manifold}
\end{equation}

To apply the continuous formulation to a point cloud, scalar
fields and differential operators must be represented at a
finite set of surface locations. We adopt a weak spatial
discretization, in which surface integration and interactions
between spatial gradients are represented by a positive definite mass matrix
$\bs{M}\in\mathbb{R}^{N\times N}$ and a symmetric positive semidefinite weak Laplacian matrix
$\bs{C}\in\mathbb{R}^{N\times N}$, respectively. The diagonal
entries of $\bs{M}$ represent the surface areas associated
with the samples, while $\bs{C}$ encodes their intrinsic
spatial coupling. These area weights allow the discretization
to account for nonuniform sampling.

Under the positive semidefinite convention adopted here,
$\bs{M}^{-1}\bs{C}$ approximates $-\Delta_{\mathcal{M}}$.
Consequently, the spatial discretization
of~\eqref{eq:diffusion_manifold} yields
\begin{equation}
    \bs{M}\frac{d\bs{u}}{ds}
    +
    \bs{C}\bs{u}
    =
    \bs{0}.
    \label{eq:semidiscrete_diffusion}
\end{equation}
Here $\bs{u}(s)\in\mathbb{R}^{N}$ collects the discrete
approximations of the scalar field at the point cloud
locations, with
$[\bs{u}(s)]_i \approx u(\bs{p}_i,s)$.
We construct $\bs{C}$ and $\bs{M}$ using the robust
point cloud Laplacian of~\citet{Sharp2020AMeshes},
implemented in the publicly available
\texttt{robust-laplacians} library.\footnote{%
\url{https://github.com/nmwsharp/robust-laplacians-py}}
The resulting matrix pair provides the common discrete
geometric representation for GP covariance construction
and ergodic exploration.

To construct the GP covariance from intrinsic surface diffusion, we use a spectral representation of the Laplace–Beltrami operator, which provides an explicit expression for the heat kernel between pairs of surface locations:
\begin{equation}
    -\Delta_{\mathcal{M}}\phi_k
    =
    \lambda_k \phi_k,
    \label{eq:lb-cont}
\end{equation}
where $\lambda_k\in\mathbb{R}_{\geq0}$ is the $k$th eigenvalue and
$\phi_k:\mathcal{M}\rightarrow\mathbb{R}$ is the corresponding
eigenfunction. Larger eigenvalues correspond to modes with faster
spatial variations.
On the point cloud, the first $n_{\mathrm{eig}}$ discrete eigenpairs are obtained from
\begin{equation}
    \bs{C}\bs{\phi}_k
    =
    \lambda_k\bs{M}\bs{\phi}_k,
    \qquad
    k=0,\ldots,n_{\mathrm{eig}}-1,
    \label{eq:discrete_eigp}
\end{equation}
where $\bs{\phi}_k\in\mathbb{R}^{N}$ is the discrete eigenvector
associated with the $k$th eigenfunction, and
$[\bs{\phi}_k]_i \approx \phi_k(\bs{p}_i)$ denotes its $i$th
component.  The discrete eigenvectors are orthonormal with respect to
the mass matrix, satisfying
$\bs{\phi}_k^\top\bs{M}\bs{\phi}_j=\delta_{kj}$, where $\delta_{kj}$
is the Kronecker delta (i.e., equal to one for $k=j$ and zero
otherwise).  This provides the discrete counterpart of orthonormality
under surface integration.  Thus, the spectral representation used for
covariance construction is derived directly from the matrix pair
governing surface diffusion.

Expanding the initial field $u_0$ in these eigenfunctions yields the spectral solution:
\begin{equation}
    u(\bs{x},s)
    =
    \sum_{k=0}^{\infty}
    \exp(-s\lambda_k)
    \left\langle u_0,\phi_k\right\rangle_{\mathcal{M}}
    \phi_k(\bs{x}).
    \label{eq:heat_spectral}
\end{equation}
Thus, diffusion attenuates the Laplace--Beltrami modes according to their intrinsic spatial frequency. The associated heat kernel is characterized by the same eigenfunctions, with each mode weighted by $\exp(-s\lambda_k)$.

\subsection{GP-Based Elasticity Mapping}
\label{sec:surface_gp}

The spatial elasticity distribution is modeled as a time-invariant scalar field
$f:\mathcal{M}\rightarrow\mathbb{R}$ defined over the explored surface. Although $f$ is not directly observable, pointwise estimates
$y_i \approx f(\bs{x}_i)$ are obtained at queried surface locations
$\bs{x}_i \in \mathcal{M}$ using the viscoelastic estimation procedure described in \autoref{sec:ukf}. The collected observations
$\bs{X}=[\bs{x}_1^\top,\ldots,\bs{x}_n^\top]^\top$ and
$\bs{y}=[y_1,\ldots,y_n]^\top$ are then used to condition the GP posterior. The GP prior mean in \eqref{eq:gp_mean} is set to the constant
\begin{equation}
    m(\mathbf{x}) = 1.1\,k_{\min}(t),
    \label{eq:prior_mean}
\end{equation}
where $k_{\min}(t)$ is the minimum elasticity estimate acquired up to
time $t$, recomputed at each GP update. Since stiff inclusions appear
as positive deviations from a softer background, this choice places
the prior close to the background level. Far from the measurements,
the posterior mean therefore reverts to a value below the surface
average.

To model smooth spatial correlations intrinsically over $\mathcal{M}$, a natural
choice would be to replace the Euclidean distance in the squared-exponential
kernel with the geodesic distance, i.e.\ the norm of the logarithmic map,
$k(\bs{x},\bs{x}') = \sigma_f^2 \exp\!\big(-\|\mathrm{Log}_{\bs{x}}(\bs{x}')\|^2 / 2l^2\big)$.
However, such geodesic exponential kernels may fail to be positive semidefinite: on curved manifolds they can produce indefinite covariance matrices
for some length scales and point configurations~\citep{Feragen2015GeodesicConflict},
which invalidates the GP posterior. Instead, the heat kernel associated with
the Laplace--Beltrami operator is used as a Riemannian analogue of the
squared-exponential covariance kernel~\citep{Borovitskiy2020MaternManifolds}. Using the eigenpairs introduced in~\eqref{eq:lb-cont}, the kernel can be expressed spectrally as
\begin{equation}
    k(\bs{x},\bs{x}')
    =
    \frac{\sigma_f^2}{Z_c}
    \sum_{k=0}^{\infty}
    \exp\!\left(-\frac{l^2}{2}\lambda_k\right)
    \phi_k(\bs{x})\phi_k(\bs{x}'),
    \label{eq:heat_kernel_spectral}
\end{equation}
where $l$ is the GP length scale, $\sigma_f^2$ is the prior variance of the elasticity field, and $Z_c$ is a normalization constant. Since $\Delta_{\mathcal{M}}$ depends only on the Riemannian metric, the resulting covariance depends on the intrinsic surface geometry rather than on Euclidean proximity in the ambient space.

On the point cloud, the continuous expansion in~\eqref{eq:heat_kernel_spectral}
is replaced by a finite sum over the $n_{\mathcal{P}}$ discrete eigenpairs
of~\eqref{eq:discrete_eigp}, giving the covariance between locations
$\bs{p}_i$ and $\bs{p}_j$:
\begin{equation}
 k_{ij}
 =
 \frac{\sigma_f^2}{Z_d}
 \sum_{k=0}^{n_{\mathcal{P}}-1}
 \exp\!\left(-\frac{l^2}{2}\lambda_k\right)
 [\bs{\phi}_k]_i[\bs{\phi}_k]_j,
 \label{eq:discrete_kernel}
\end{equation}
where the normalization constant $Z_d$ is chosen such that the mass-weighted
average prior variance over the surface equals $\sigma_f^2$. Since the
eigenvectors are $\mathbf{M}$-orthonormal, \eqref{eq:discrete_kernel} is
exactly the heat kernel of the discrete operator pair $(\mathbf{C},\mathbf{M})$,
which in matrix form reads
$\mathbf{K} = \frac{\sigma_f^2}{Z_d}\exp\!\left(-\frac{l^2}{2}\mathbf{M}^{-1}\mathbf{C}\right)\mathbf{M}^{-1}$.
Evaluating it, however, requires the full eigendecomposition of an
$n_{\mathcal{P}} \times n_{\mathcal{P}}$ generalized eigenvalue problem, which becomes expensive for
dense point clouds. Because the weights $\exp(-l^2\lambda_k/2)$ decay rapidly
with $\lambda_k$, high-frequency modes contribute little to the covariance,
and the sum can be restricted to the $n_{\mathrm{eig}} \ll n_{\mathcal{P}}$ eigenpairs with
the smallest eigenvalues:
\begin{equation}
 k_{ij}
 \approx
 \frac{\sigma_f^2}{Z_d}
 \sum_{k=0}^{n_{\mathrm{eig}}-1}
 \exp\!\left(-\frac{l^2}{2}\lambda_k\right)
 [\bs{\phi}_k]_i[\bs{\phi}_k]_j,
 \label{eq:discrete_kernel_truncated}
\end{equation}
with $Z_d$ computed for the truncated sum. These eigenpairs are obtained with a sparse eigensolver, which substantially
reduces the cost of the offline pre-processing stage. The truncation is
accurate when the weight of the last retained mode is small; its effect on reconstruction accuracy is assessed empirically
in~\autoref{sec:gpr_pointclouds}.
Note that, unlike geodesic-distance kernels, both~\eqref{eq:discrete_kernel}
and its truncation~\eqref{eq:discrete_kernel_truncated} are positive
semidefinite by construction, being sums of rank-one matrices
$\bs{\phi}_k\bs{\phi}_k^\top$ with strictly positive weights. Since the
truncated kernel has rank at most $n_{\mathrm{eig}}$, the inverse
in~\eqref{eq:gp_mean} is well defined owing to the noise term
$\sigma_n^2\mathbf{I}$.

Under the fixed reference approximation introduced
in~\autoref{sec:surface_geometry}, the eigenpairs
in~\eqref{eq:discrete_eigp} and the pairwise kernel values
in~\eqref{eq:discrete_kernel_truncated} can be precomputed during
initialization. Specifically, $k_{ij}$ is evaluated for every pair of points
$(i,j)$ in the reference point cloud, producing a lookup table that encodes
spatial correlations along the surface. During online execution, each
contact location along the exploration trajectory is associated with the
closest point in the reference point cloud, so that the covariance matrices
in~\eqref{eq:gp_mean} and~\eqref{eq:gp_cov} are assembled directly from the
precomputed $k_{ij}$ values.

During exploration, the GP is conditioned on the elasticity estimates
collected along the trajectory and evaluated at all point cloud vertices,
yielding, through~\eqref{eq:gp_mean} and~\eqref{eq:gp_cov}, the posterior
mean $\mu$ and standard deviation $\sigma$ of the elasticity field over the
surface. The surface geometry enters the model only through the kernel,
which determines how each measurement propagates along the surface, whereas
the measured elasticity enters only through the observations.

\subsection{EID-Driven Ergodic Exploration on Point Clouds}
\label{sec:ergodic_pc}

The exploration policy builds on the HEDAC formulation of~\citet{Bilaloglu2025TactileSurfaces}, which drives the robot so that
its time-averaged coverage of the surface matches a fixed target distribution
$p(\bs{x})$. For the present
information-driven palpation problem, the target $p(\bs{x},t)$ is instead
defined online by combining the GP posterior mean $\mu$ and standard
deviation $\sigma$ of~\autoref{sec:surface_gp} into the expected
information density (EID):
\begin{equation}
\begin{split}
  p(\bs{x},t) := & \tilde{\xi}_{\mathrm{EID}}(\bs{x},t) = (1-\alpha)
  \frac{\sigma(\bs{x},t)}
  {\int_{\mathcal{M}} \sigma(\bs{x},t)\,dA} \\
  & + \alpha \frac{\max(\mu(\bs{x},t)-\bar{\mu}(t),0)}
  {\int_{\mathcal{M}} \max(\mu(\bs{x},t)-\bar{\mu}(t),0)\,dA +
    \epsilon},
\end{split}
\label{eq:eid}
\end{equation}
where $\bar{\mu}(t)$ is the area-weighted average of the posterior
mean over $\mathcal{M}$, and the integrals are evaluated on the point
cloud using the area weights in $\mathbf{M}$. As previously mentioned,
far from the measurements, the posterior mean therefore reverts to a
value below $\bar{\mu}(t)$, so unexplored regions receive no weight
from the exploitation term of the EID~\eqref{eq:eid} and spurious
stiffness peaks are avoided.
Notice that the
posterior standard deviation~\eqref{eq:gp_cov} is independent from
this choice.  Therefore, the first term in~\eqref{eq:eid} favors
regions of high posterior uncertainty; the second emphasizes positive
deviations from the current average elasticity, increasing the
relative importance of spatial stiffness peaks. Since both terms are
normalized, the coefficient $\alpha \in [0,1]$ directly sets the
balance between exploitation and exploration, while a small
$\epsilon > 0$ ensures numerical robustness when no positive deviation
is present.

Then, at each system time $t$, the current target distribution
$\tilde{\xi}_{\mathrm{EID}}(\boldsymbol{x},t)$ is compared with the
normalized coverage $\bar{c}(\bs{x},t)$
in~\autoref{sec:ergodic_control}. Their positive difference identifies
regions that have received less visitation than currently desired,
while its square assigns greater weight to larger mismatches:
\begin{equation}
    u_0(\bs{x},t)
    =
    \max\!\left(
        \tilde{\xi}_{\mathrm{EID}}(\boldsymbol{x},t)
        -
        \bar{c}(\bs{x},t),
        0
    \right)^2.
    \label{eq:hedac_source}
\end{equation}
The resulting field $u_0$, sampled at the point cloud vertices as
$\bs{u}_0(t)$, is diffused according to the semi-discrete heat equation in~\eqref{eq:semidiscrete_diffusion}. Taking $\bs{u}(0)=\bs{u}_0(t)$ as
initial condition, the system is integrated in the diffusion variable $s$
up to a fixed diffusion time $s=\tau$, which is independent of the physical exploration time $t$: at
each system time, a new diffusion is computed from the current source
$\bs{u}_0(t)$, and its result is denoted $\bs{u}_\tau(t)$. Diffusion propagates the influence of underexplored regions across the surface, producing a smooth potential field whose gradient provides the exploration direction.

Using the positive-semidefinite Laplacian formulation, $\bs{C}$ discretizes $-\Delta_{\mathcal{M}}$, while $\bs{M}$ is the corresponding mass matrix.
Applying one backward-Euler step on \eqref{eq:semidiscrete_diffusion} of duration $\tau$ from the initial field $\bs{u}_0(t)$ gives
\begin{equation}
    \bs{M}
    \frac{\bs{u}_{\tau}(t)-\bs{u}_0(t)}{\tau}
    +
    \bs{C}\bs{u}_{\tau}(t)
    =
    \bs{0},
\end{equation}
which can be rearranged as the sparse linear system
\begin{equation}
    \bigl(\bs{M}+\tau\bs{C}\bigr)\bs{u}_{\tau}(t)
    =
    \bs{M}\bs{u}_0(t).
    \label{eq:splu}
\end{equation}
Since the mass matrix $\mathbf{M}$ is positive definite and the weak Laplacian $\mathbf{C}$ is positive semidefinite, $\mathbf{M}+\tau\mathbf{C}$ is positive definite for $\tau>0$, ensuring a unique solution.
Since $\bs{M}+\tau\bs{C}$ is sparse, $\bs{u}_{\tau}(t)$ is computed using a sparse LU factorization followed by forward and backward substitution, without explicitly computing a matrix inverse. Because the surface geometry and $\tau$ remain fixed during exploration, the factorization can be reused at each controller update, while only the right-hand side $\bs{M}\bs{u}_0(t)$ changes.

Solving~\eqref{eq:splu} with the full discrete operators $(\mathbf{C},\mathbf{M})$, avoids truncating the Laplace--Beltrami spectrum, while diffusion itself regularizes the initial field. The diffusion time $\tau$
is scaled with the point cloud sampling resolution as $\tau=\chi h^2$,
where $h$ is the mean distance between each point of $\mathcal{P}$ and its
nearest neighbor and $\chi>0$ is a dimensionless hyperparameter.
Increasing $\chi$ produces a smoother and more spatially distributed
potential field, whereas its reduction preserves more localized
spatial variations.

The spatial gradient of the diffused potential field at the continuous end-effector location $\bs{x}_{EE}\in\mathbb{R}^3$ serves as the control command for the acceleration-controlled exploration agent, which generates the reference trajectory for the interaction controller:
\begin{equation}
    \ddot{\bs{x}}_{EE}(t)
    =
    \nabla_{\bs{x}} u_{\tau}(\bs{x}_{EE};t).
    \label{eq:ee_agent}
\end{equation}
The gradient is projected onto the local tangent plane of the surface, so that the agent moves along the surface while motion along the normal is left to force regulation.
Since $\bs{u}_{\tau}(t)$ is defined only at the point cloud vertices, evaluating the gradient at an arbitrary off-vertex position requires a continuous local representation. Let $\mathcal{N}(\bs{x}_{EE})\subset\mathcal{P}$ denote the $n_{\mathcal{N}}$-nearest spatial neighbors of $\bs{x}_{EE}$. A local approximation $\hat{u}(\bs{x};t)$ is constructed from the corresponding values of $\bs{u}_{\tau}(t)$ by fitting a third-degree polynomial via least squares. The continuous gradient is then evaluated analytically at the end-effector position:
\begin{equation}
    \nabla_{\bs{x}}
    u_{\tau}(\bs{x}_{EE};t)
    \approx
    \left.
    \nabla_{\bs{x}}
    \hat{u}(\bs{x};t)
    \right|_{\bs{x}=\bs{x}_{EE}}.
    \label{eq:local_gradient}
\end{equation}
This provides a smooth differentiable vector field between point cloud samples without requiring an explicit surface reconstruction.

\paragraph{Decoupling of Model Update and Planning}
GP regression and ergodic planning operate asynchronously and at different rates. At the planning rate, the accumulated visitation and the potential field are updated using the most recent EID. Meanwhile, new elasticity measurements are incorporated into the GP as they become available. Once a GP update is completed, the resulting EID replaces the current target distribution at the next planning update, without interrupting the exploration motion.

\begin{figure*}[t]
    \centering
    \subfloat[Folded dimple with Gaussian-shaped informative region.]{\label{1a}%
            \includegraphics[width=0.33\textwidth]{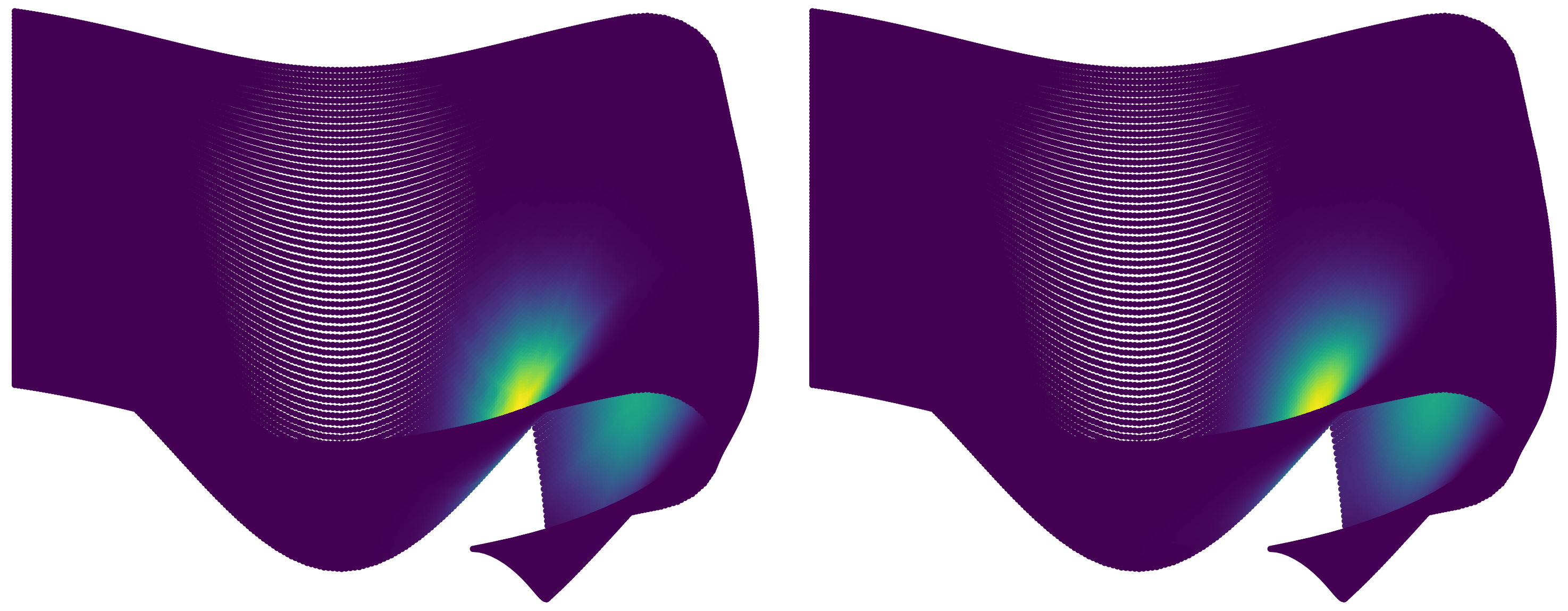}}
    \hfill
    \subfloat[Stanford bunny with X-shaped informative region.]{\label{1b}%
            \includegraphics[width=0.33\textwidth]{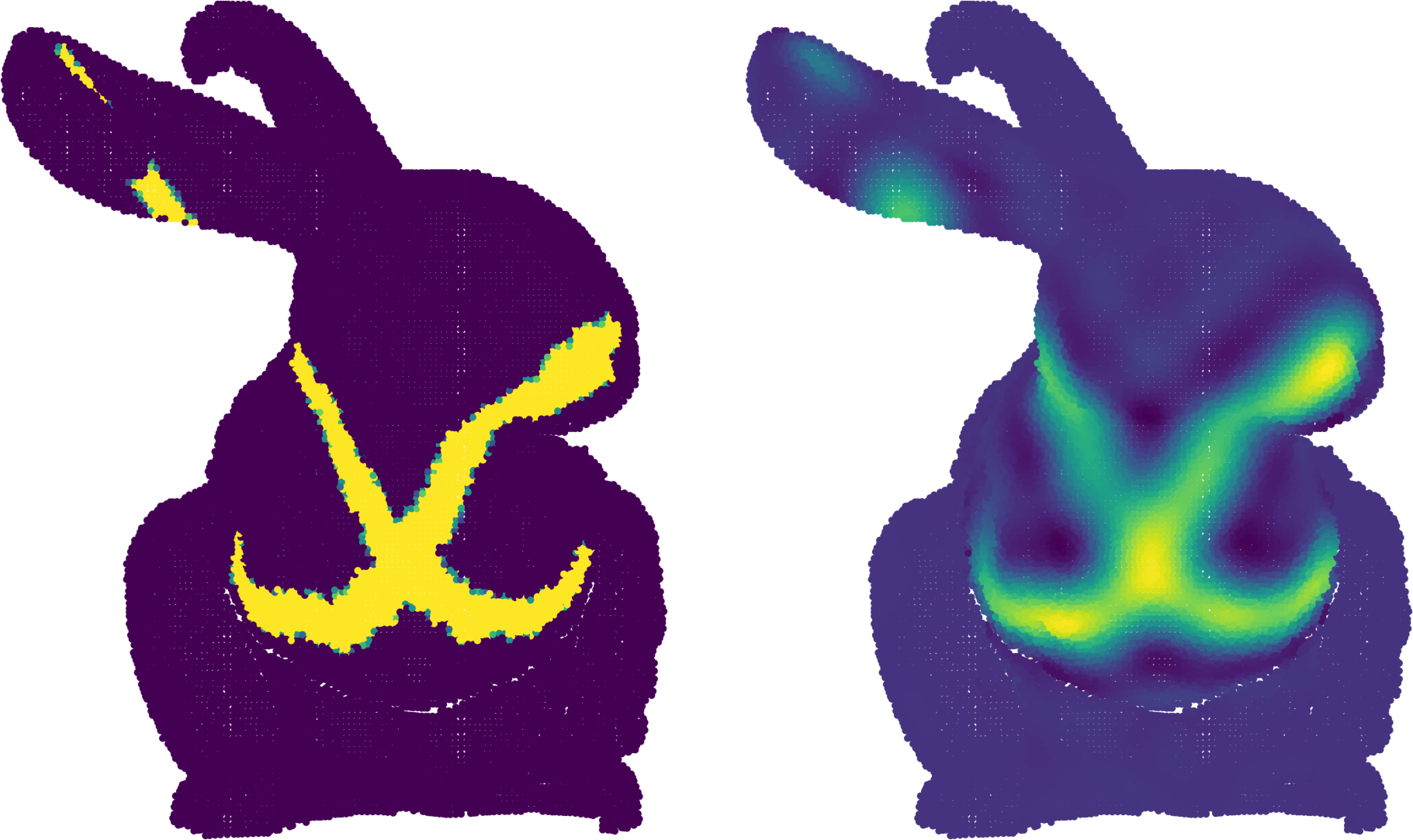}}
    \hfill
    \subfloat[Plate with multiple informative regions with different shapes.]{\label{1c}%
            \includegraphics[width=0.33\textwidth]{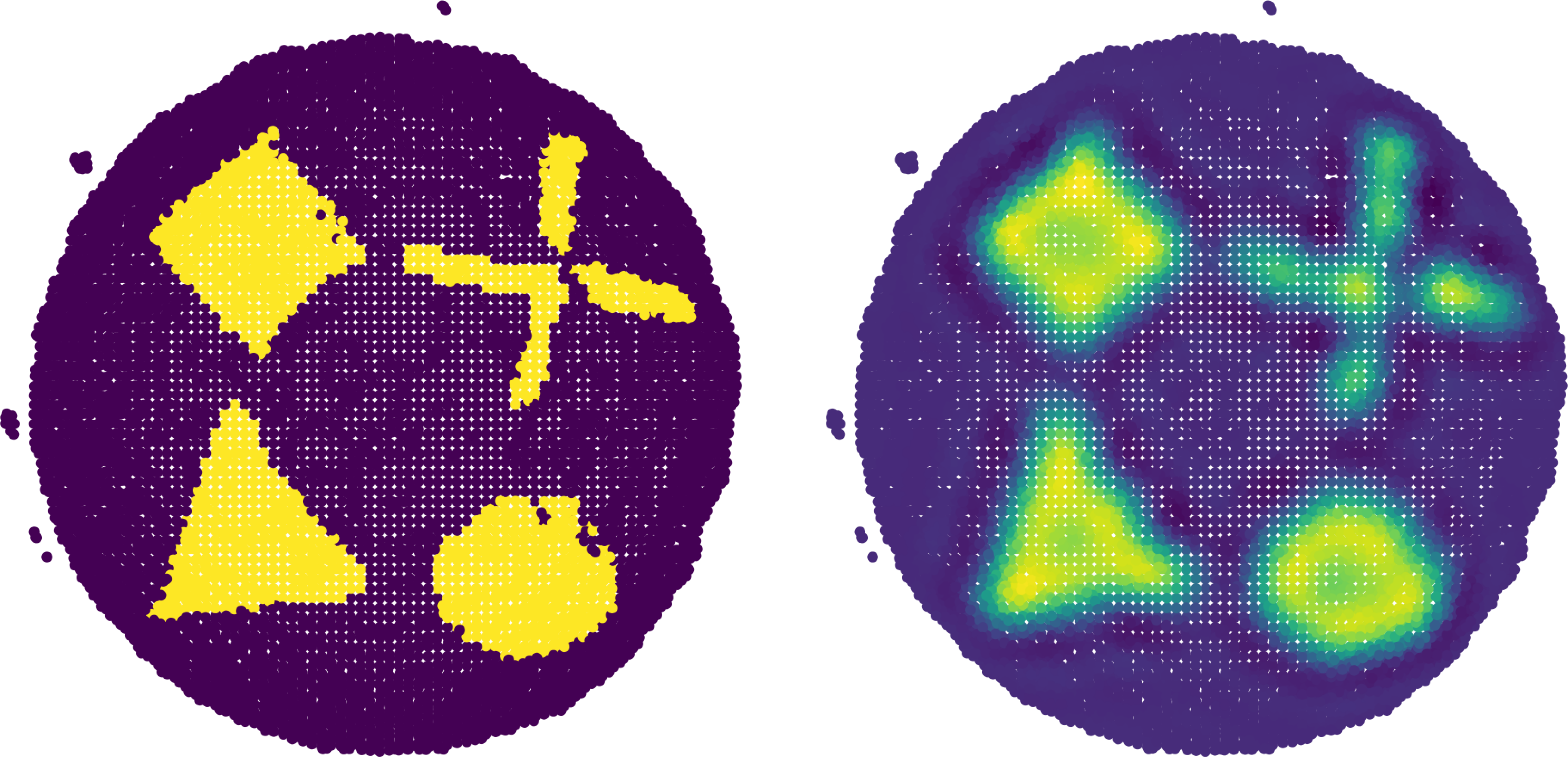}}
    \hfill
    \caption{
    On the left, folded-dimple, Bunny, and Plate point clouds are used as domains for the synthetic target information
distributions.
    On the right of each subplot, the reconstruction was obtained using 30\% of the points normally sampled from the point cloud. Blue denotes regions of low information density, whereas yellow indicates regions of high information density.
    }
    \label{fig:gpr_test}
\end{figure*}

\section{Simulation Results on Real Point Clouds}
In this section, we evaluate the search and coverage
performance of the proposed ergodic exploration framework
in simulation. The experiments use both real surface
geometries acquired using an Intel RealSense D405 and a
synthetically generated folded-dimple geometry, with synthetic
information distributions defined over these surfaces to
emulate spatially varying stiffness patterns. The physical
palpation process is replaced by direct sampling of the
prescribed synthetic distribution (used as ground truth),
allowing controlled assessment of exploration behavior
independently of sensing noise and contact dynamics.
For visualization purposes, the point clouds are colored
according to the target information density, with blue
indicating low values and yellow indicating high values.

\subsection{GPR on Point Clouds}
\label{sec:gpr_pointclouds}
The Gaussian Process Regression (GPR) is the most computationally demanding
component of the framework. Since exact GP inference scales cubically
with the number of observations, maintaining synchronous model updates
at high control rates can become prohibitive as the map grows. The
analysis in this subsection uses the plate point cloud
of~\autoref{fig:gpr_test}c, which contains $n_\mathcal{P}=7235$ points
after voxel downsampling.

The impact of the number of eigenfunctions and regression points is
evaluated via the root mean square error (RMSE) between the
reconstructed distribution and the ground-truth synthetic stiffness
field, whose values are encoded in the point-cloud color map. From now
on, the RMSE is defined as
\begin{equation}
\mathrm{RMSE}(t)
=
\sqrt{
\frac{1}{n_\mathcal{P}}
\left\|
\bs{\mu}_{(t)} - \bs{\mu}_{\mathrm{GT}}
\right\|_2^2
},
\label{eq:rmse}
\end{equation}
where $\bs{\mu}_(t)$ is the mean value of the GP regression computed over the point cloud at time $t$. $\bs{\mu}_{\mathrm{GT}}$ is the ground truth and $n_\mathcal{P}$ is the number of points of the point cloud.
\begin{figure}[t]
    \centering
    \includegraphics[width=0.9\linewidth, trim={2mm 2mm 0mm 0mm}, clip]{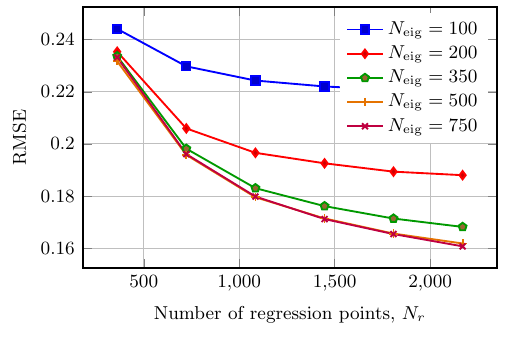}
    \caption{Effect of increasing the number of regression points for different fixed numbers of eigenfunctions. Note that the curve with $n_{eig} = 750$ overlaps the one with $n_{eig} = 500$.}
    \label{fig:eig_eff}
\end{figure}
\autoref{fig:eig_eff} shows that increasing the number of eigenfunctions of~\eqref{eq:discrete_kernel_truncated} leads to a reduction in RMSE up to $n_{eig} = 500$, beyond which no significant improvement is observed. Moreover, the RMSE decreases steeply up to $1000$ regression points, after which performance saturates. As expected, the kernel computation time increases markedly with the number of eigenfunctions, reaching approximately \SI{5}{\second} for $n_{eig} = 500$, \SI{19}{\second} for $n_{eig} = 750$, and \SI{29}{\second} for $n_{eig} = 1000$. Since the kernel matrix $k_{i,j}$ is precomputed offline during this phase, this cost only affects the initialization stage and does not impact the online execution of the ergodic planner. Considering the trade-off between reconstruction accuracy and computational overhead, $n_{eig} = 500$ is adopted for all the considered point clouds, except for the rabbit geometry. The latter represents a particularly challenging case, as its point-cloud representation consists of four spatially separated clusters, with gaps ranging from a few millimeters to approximately \SI{1}{\centi\meter}. To accommodate this more challenging representation, the number of eigenfunctions is increased to $n_{eig} = 2500$. This case demonstrates the applicability of the proposed approach to point clouds with substantial spatial discontinuities, at the expense of increased offline computational cost.
\begin{table}[t]
\centering
\caption{Online prediction time and update rate for increasing size of the
regression set $N_r$ ($n_\mathcal{P}=7235$ points,
$n_{eig}=500$).}
\label{tab:pred_time}
\setlength{\tabcolsep}{4pt}
\begin{tabular}{ccc}
\toprule
$\bm{N_r}$ & \textbf{Prediction Time} (s) & \textbf{Update Rate} (Hz)\\
\midrule
300 & 0.0062 & 161.8 \\
600 & 0.0098 & 102.4 \\
900 & 0.0131 & 76.2 \\
1200 & 0.0204 & 49.1 \\
1800 & 0.0323 & 31.0 \\
2000 & 0.0367 & 27.3 \\
\bottomrule
\end{tabular}
\end{table}

Computational benchmarks were performed on a PC equipped with AMD Ryzen 9 5900X 12-Core, 32GB of RAM, and NVIDIA GeForce RTX 4070 running Ubuntu 24.04.
\autoref{tab:pred_time} reports the prediction time and update rate for different numbers of regression points with $n_{eig} = 500$. Consistent with the theoretical scaling of GP regression, increasing the number of regression points improves reconstruction accuracy at the expense of higher prediction time. The update rate varies from approximately \SI{160}{\hertz} using $N_r=300$ to approx \SI{27}{\hertz} using $N_r=2000$. In the proposed parallel implementation, GP prediction updates are executed asynchronously and do not interfere with the ergodic planning loop. This further highlights the suitability of ergodic exploration for computationally intensive estimation tasks, as it enables continuous motion generation while the underlying model is refined in the background. In particular, when measurements are acquired at a rate below \SI{20}{\hertz} and the number of regression points is below 2000, each GP update completes before the next sample arrives, so the prediction always conditions on the most recent available data and no measurement is delayed to a subsequent update cycle.
\autoref{fig:gpr_test} presents a qualitative comparison between the original point clouds and the corresponding reconstructions. For each object, the original point cloud is shown together with the reconstructed surface obtained via GPR using \SI{30}{\percent} of the available points.

\subsection{Effect of the Surface Metric on Regression}
\label{sec:kernel_ablation}
This experiment isolates the contribution of the manifold-aware
covariance model by comparing the proposed GPR based on the intrinsic
heat kernel~\eqref{eq:heat_kernel_spectral} with GPR using a
squared-exponential kernel (RBF) using the RMSE defined
in~\eqref{eq:rmse}. All other components of the regression are kept
fixed, so that differences in reconstruction accuracy can be only
attributed to the definition of the spatial correlation. The two
distance notions are similar on locally flat surface regions, but may
differ substantially when points that are close in the ambient space
are distant along the surface, e.g., across folds, holes or
disconnected regions.  The comparison therefore evaluates how the
benefit of the intrinsic kernel depends on the surface geometry and
when the additional eigendecomposition is justified.

Both models are evaluated on the acquired Bunny and Plate
point clouds and on a synthetic folded-dimple geometry,
using the same target fields, vertex sets, and length scale,
$l = 5\,\mathrm{mm}$.
Each configuration is evaluated over ten random nested
sets of regression points.

\begin{table}[t]
\centering
\caption{Intrinsic (heat) versus Euclidean (RBF) covariance under
identical conditions: same point clouds, same target fields, lengthscale
$l=5\,\mathrm{mm}$ for the intrinsic
construction. For each object and training fraction, both kernels are
conditioned on the same vertices, so the comparison is paired. Test RMSE
is evaluated on the held-out vertices and reported as mean $\pm$
standard deviation over ten random nested regression-point sets.}
\label{tab:kernel_comparison_objects}
\footnotesize
\setlength{\tabcolsep}{4pt}
\begin{tabular}{llccc}
\toprule
Geometry & (\%) & Euclidean & Heat kernel & Reduction \\
\midrule
\multirow{4}{*}{Bunny}
 & 5 & $\mathbf{0.201 \pm 0.015}$ & $0.208 \pm 0.015$ & -3.7\% \\
 & 10 & $\mathbf{0.162 \pm 0.007}$ & $0.169 \pm 0.008$ & -3.9\% \\
 & 15 & $\mathbf{0.140 \pm 0.006}$ & $0.145 \pm 0.006$ & -3.6\% \\
 & 20 & $\mathbf{0.130 \pm 0.006}$ & $0.134 \pm 0.007$ & -3.1\% \\
\midrule
\multirow{4}{*}{Plate}
 & 5 & $0.309 \pm 0.017$ & $\mathbf{0.298 \pm 0.018}$ & 3.5\% \\
 & 10 & $0.243 \pm 0.015$ & $\mathbf{0.231 \pm 0.015}$ & 4.9\% \\
 & 15 & $0.206 \pm 0.007$ & $\mathbf{0.199 \pm 0.006}$ & 3.6\% \\
 & 20 & $0.184 \pm 0.007$ & $\mathbf{0.181 \pm 0.007}$ & 1.4\% \\
\midrule
\multirow{4}{*}{\shortstack{Folded\\Dimple}}
 & 5 & $\mathbf{0.063 \pm 0.013}$ & $0.065 \pm 0.023$ & -4.1\% \\
 & 10 & $0.039 \pm 0.007$ & $\mathbf{0.034 \pm 0.007}$ & 11.1\% \\
 & 15 & $0.030 \pm 0.004$ & $\mathbf{0.025 \pm 0.003}$ & 14.3\% \\
 & 20 & $0.024 \pm 0.002$ & $\mathbf{0.021 \pm 0.002}$ & 11.5\% \\
\bottomrule
\end{tabular}
\end{table}

The results in~\autoref{tab:kernel_comparison_objects} demonstrate the benefits of intrinsic covariance for elastic field reconstruction across different surface geometries. On the Bunny, the two kernels achieve comparable accuracy, with the Euclidean kernel yielding slightly lower RMSE (3.1\%--3.9\%). On the Plate, the intrinsic kernel consistently improves reconstruction accuracy, with RMSE reductions ranging from 1.4\% to 4.9\%. More substantial improvements are observed on the Folded Dimple, where the intrinsic kernel reduces RMSE by 11.1\%, 14.3\%, and 11.5\% for training fractions of 10\%, 15\%, and 20\%, respectively. These results highlight that the advantage of intrinsic covariance becomes particularly pronounced when the surface geometry introduces discrepancies between Euclidean proximity and intrinsic connectivity. In the Folded Dimple, points that are close in $\mathbb{R}^3$ may be distant along the surface, causing Euclidean covariance to introduce spurious correlations across geometrically separated regions. The heat kernel instead preserves intrinsic surface relationships, improving reconstruction accuracy without requiring additional measurements. Importantly, across the evaluated geometries, these improvements are achieved without substantial reconstruction accuracy penalties in cases where intrinsic modeling provides limited benefits. Therefore, despite the additional offline preprocessing required to construct the kernel, intrinsic covariance represents a robust choice for elastic mapping on curved surfaces, offering considerable advantages in geometrically challenging configurations while maintaining comparable performance elsewhere. This property is particularly relevant for anatomical surfaces characterized by folds, concavities, and closely spaced regions.

\subsection{Effect of the Update Interval}

\begin{figure}[t]
    \centering
    \includegraphics[width=1\linewidth]{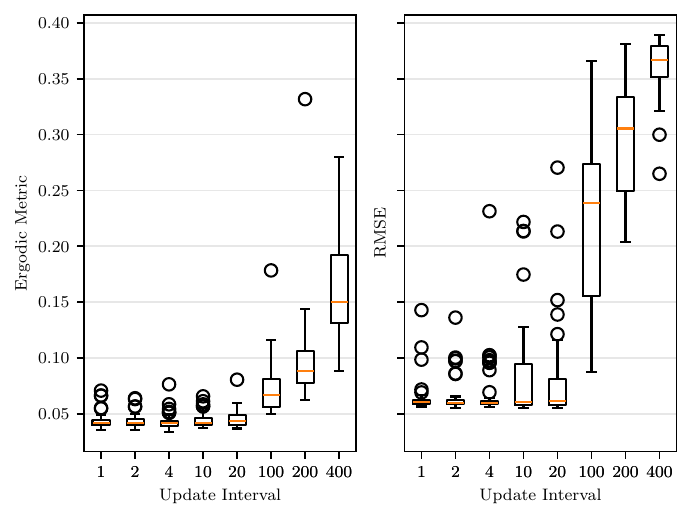}
    \caption{Effect of the update interval on performance. Boxplots show the ergodic metric (left) and RMSE (right) for increasing numbers of samples between consecutive GP updates. Frequent GP updates keep the trajectory aligned with the evolving EID, resulting in low ergodic metric and RMSE.}
    \label{fig:update_interval}
\end{figure}

Before examining the search behavior and the influence of the exploration coefficient, we first analyze the effect of the update interval on overall performance.
The update interval denotes the number of samples collected before the
GP is updated via regression with the newly acquired data and
determines how frequently the estimated distribution is updated using
newly acquired measurements, therefore directly affecting the
responsiveness of the search strategy. \autoref{fig:update_interval}
reports the distribution of the ergodic metric and the reconstruction
error (RMSE) at the end of the exploration time for increasing update
intervals over multiple runs.
\begin{figure*}[t]
    \centering
    \begin{tikzpicture}
        \node (img) {
            \begin{minipage}{\linewidth}
                \centering
                \includegraphics[]{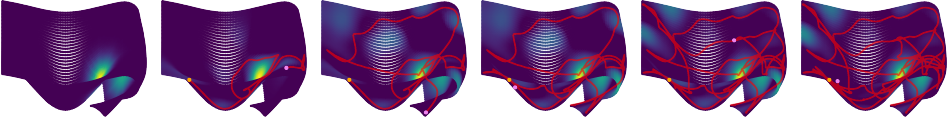}\\
                \includegraphics[]{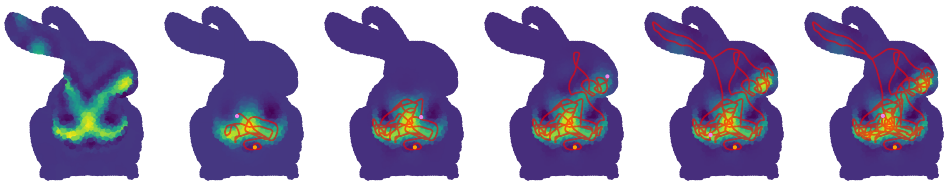}\\
                \includegraphics[]{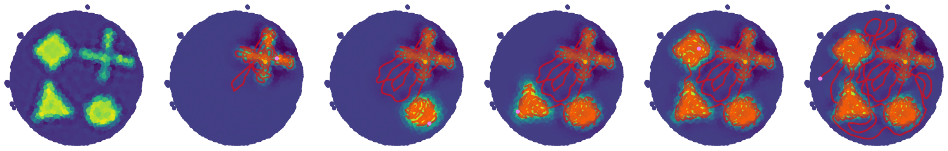}
            \end{minipage}
        };

        \node[rotate=90, anchor=east] at ($(img.north west)+(0.2cm,-0.0cm)$) {\footnotesize Folded Dimple};
        \node[rotate=90, anchor=east] at ($(img.north west)+(0.2cm,-3.5cm)$) {\footnotesize Bunny};
        \node[rotate=90, anchor=east] at ($(img.north west)+(0.2cm,-6.5cm)$) {\footnotesize Plate};

        \node[anchor=west] at ($(img.north west)+(0.6cm,+0.5cm)$) {\footnotesize \textbf{Original Distribution}};
        \node[anchor=west] at ($(img.north)+(0cm,+0.5cm)$) {\footnotesize \textbf{Time Evolution}};
    \end{tikzpicture}
    \caption{
    Time evolution of the ergodic search with a fixed exploration coefficient $\alpha = 0.6$ for different point-cloud shapes.
    The left column shows the underlying target distributions, while the right columns illustrate the temporal evolution of the search at uniformly spaced time intervals, including updates of the estimated distribution and the agent’s trajectory (red).
    }
    \label{fig:shapes_search}
\end{figure*}
To define the ergodic metric $em(t)\in\mathbb{R}$, let $C_i(t)$ denote the discrete accumulated coverage at point-cloud vertex $\mathbf{x}_i$ along the executed trajectory, introduced in \autoref{sec:ergodic_control}. In the experimental implementation, it is
updated along the executed trajectory as
\begin{equation}
C_i(t+\Delta t)=C_i(t)+
\exp\!\left[-R^{-2}\|\mathbf{x}_i-\mathbf{x}_{\mathrm{EE}}(t)\|_2^2\right],
\label{eq:coverage_update}
\end{equation}
where $R$ is the indenter radius, adopted as the characteristic
length of the empirical Gaussian footprint. The accumulated visitation and
current target densities are normalized as
\begin{equation}
\rho_i(t)=\frac{C_i(t)}{\sum_j C_j(t)+\varepsilon},
\qquad
\rho_i^\ast(t)=
\frac{\tilde{\xi}_{\mathrm{EID},i}(t)}
{\sum_j\tilde{\xi}_{\mathrm{EID},j}(t)+\varepsilon},
\label{eq:coverage_densities}
\end{equation}
with $\varepsilon=10^{-12}$. Their normalized discrepancy is
\begin{equation}
\mathrm{em}(t)=
\frac{
\frac{1}{N}\sum_{i=1}^{N}
|\rho_i^\ast(t)-\rho_i(t)|
}{\mathrm{em}(0)},
\label{eq:coverage_metric}
\end{equation}
where $\mathrm{em}(0)$ is the initial discrepancy.  For small update
intervals (high-frequency updates), both the ergodic metric and the
RMSE remain consistently low, indicating that frequent updates enable
the agent to closely match the target distribution while maintaining
an accurate estimate of the underlying stiffness field (see
\autoref{fig:update_interval}). In this regime, newly acquired
information is rapidly incorporated, resulting in stable convergence
and limited variability across trials. As the update interval
increases (lower frequency updates), performance progressively
degrades. Beyond 100 samples for each update interval, both metrics
exhibit a marked increase in median value and dispersion. Large update
intervals delay the incorporation of measurements, causing the agent
to rely on outdated estimates and leading to suboptimal coverage of
salient regions. This behavior is reflected in the sharp increase of
the ergodic metric, as well as in the substantial growth of the
reconstruction error.  Also, these results highlight a trade-off
between computational efficiency and performance. While larger update
intervals reduce computational overhead, they significantly degrade
both exploration quality and estimation accuracy.

\subsection{Ergodic Search on Point-Clouds}
\label{ssec:ergodic_search_sim}
Having characterized the influence of the main GP design parameters, we
now evaluate
the proposed algorithm in a simplified setting, focusing on its search
and coverage characteristics under a fixed exploration coefficient
$\alpha$,~\eqref{eq:eid}. The
objective is to identify the location and spatial extent of regions
with high target information density (highlighted in yellow), without
any prior knowledge of their shape or location.  The same point-cloud
shapes in~\autoref{fig:gpr_test}, previously used for validating the
heat-kernel-based reconstruction, are employed here as target
information distributions.
\autoref{fig:shapes_search} shows the time evolution of the ergodic
search for different shapes with a fixed intermediate value of
$\alpha = 0.6$. The leftmost column reports the ground-truth target
distributions, whereas the remaining columns illustrate successive
stages of the search, including online updates of the estimated
distribution and the agent trajectory (red). At the beginning, the
absence of information leads to exploratory behavior covering the
workspace. As measurements are incorporated, the estimated
distribution progressively reveals the salient regions, and the
trajectory increasingly concentrates around these areas. Once a region
is sufficiently explored, the agent is driven towards unexplored
areas, enabling the discovery of additional salient regions. This
behavior is particularly evident for the Plate point cloud, where
distinct regions are visited sequentially before a final phase
dominated by residual exploration.

\begin{figure*}[t!]
    \centering

\begin{tikzpicture}

\def\ColSep{3mm}
\def\RowSep{0mm}

\node[anchor=north west, inner sep=0, outer sep=0, xshift=10mm] (LeftBlock) {
  \begin{minipage}[t]{0.23\linewidth}
    \centering
    \includegraphics[width=0.85\linewidth]{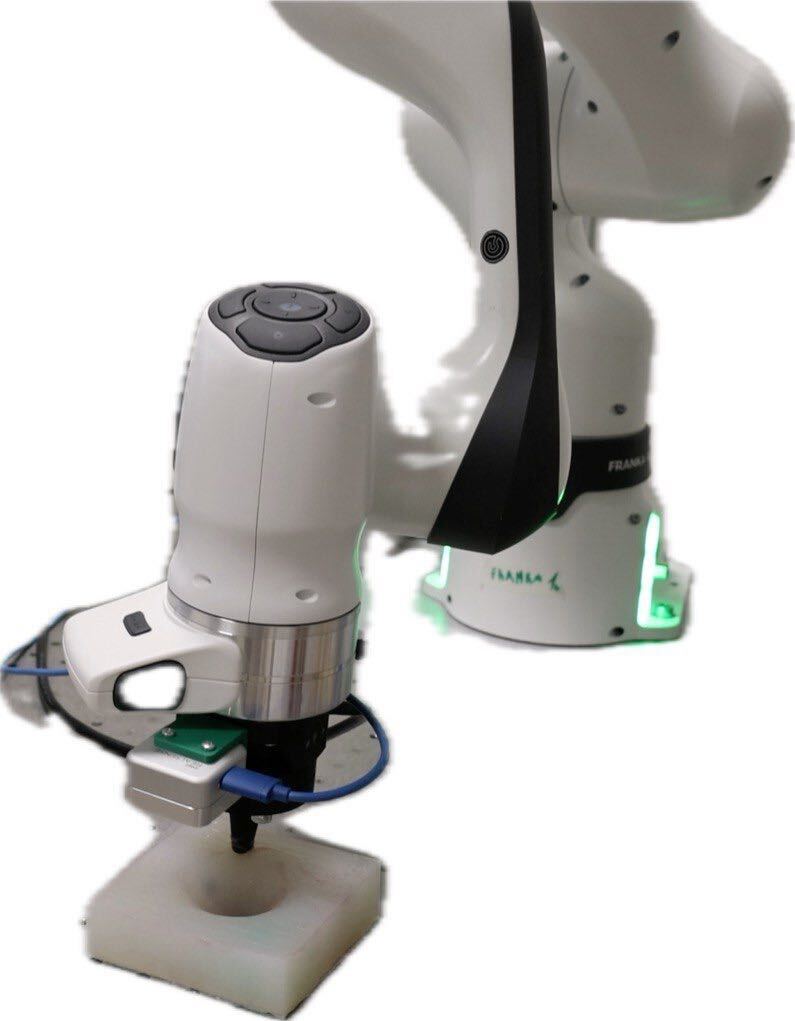}\par\vspace{3mm}
    {\footnotesize\textbf{Original Distribution}}\par\vspace{2mm}

    \begin{tikzpicture}[inner sep=0, outer sep=0]
      \node (OD) {\includegraphics[width=0.7\linewidth, trim=0cm 0cm 13.5cm 0cm, clip]
        {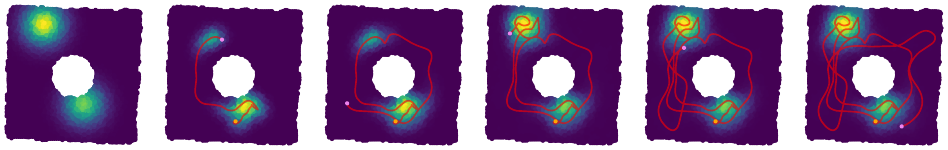}};
    \end{tikzpicture}
  \end{minipage}
};

\node[anchor=north west, inner sep=0, outer sep=0, right=\ColSep of LeftBlock.east] (RightBlock) {
  \begin{minipage}[t]{0.63\linewidth}
    \centering
    {\footnotesize\textbf{Time Evolution}}\par\vspace{2mm}

    \begin{tikzpicture}[inner sep=0, outer sep=0]

      \node (R1) {\includegraphics[width=\linewidth, trim=2.7cm 0cm 0cm 0cm, clip]
        {images/simulations/distribution_evolution_sil_0_2.pdf}};
      \node[rotate=90, anchor=south] at ($(R1.west)+(-6mm,0)$) {\footnotesize $\alpha = 0.2$};

      \node[below=\RowSep of R1.south west, anchor=north west] (R2)
        {\includegraphics[width=\linewidth, trim=2.7cm 0cm 0cm 0cm, clip]
        {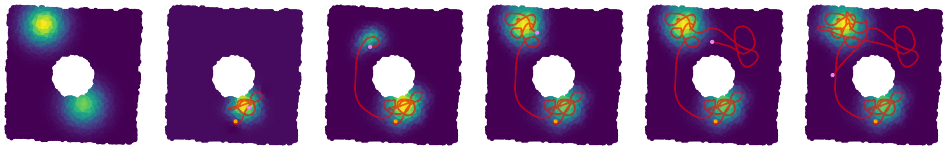}};
      \node[rotate=90, anchor=south] at ($(R2.west)+(-6mm,0)$) {\footnotesize $\alpha = 0.4$};

      \node[below=\RowSep of R2.south west, anchor=north west] (R3)
        {\includegraphics[width=\linewidth, trim=2.7cm 0cm 0cm 0cm, clip]
        {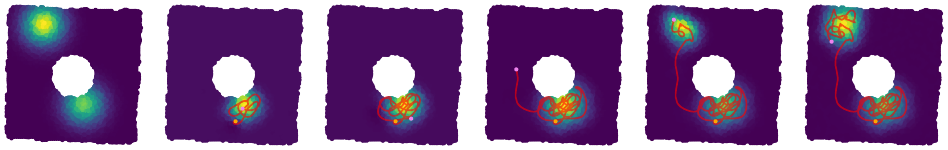}};
      \node[rotate=90, anchor=south] at ($(R3.west)+(-6mm,0)$) {\footnotesize $\alpha = 0.6$};

      \node[below=\RowSep of R3.south west, anchor=north west] (R4)
        {\includegraphics[width=\linewidth, trim=2.7cm 0cm 0cm 0cm, clip]
        {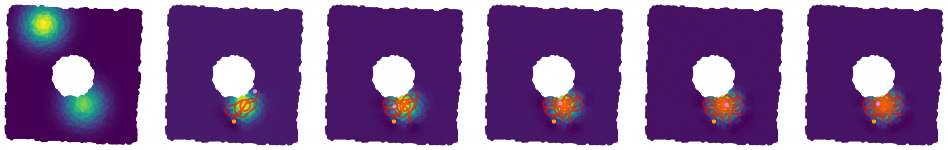}};
      \node[rotate=90, anchor=south] at ($(R4.west)+(-6mm,0)$) {\footnotesize $\alpha = 0.8$};

    \end{tikzpicture}
  \end{minipage}
};

\end{tikzpicture}

    \caption{Comparison of the time evolution of the ergodic search for different values of the exploration coefficient $\alpha$. (top left) Gaussian Dimple phantom used to generate the pointcloud; (bottom left) underlying target distribution; (right) search progress over time displayed at uniformly spaced time intervals. Each successive column corresponds to the same fixed simulation time increment, and all sequences are shown up to a final time of $\SI{150}{\second}$. The plots include the updated distribution estimate and the agent’s trajectory (red). Top view of the sample; the dimple is truncated for visualization, appearing as an interior hole.}
    \label{fig:alpha_eff}
\end{figure*}

\subsection{Balancing Exploitation and Exploration}
We now analyze the role of the exploration coefficient $\alpha$, which
regulates the trade-off between exploration and exploitation by
weighting the relative emphasis on global coverage versus local
refinement of the estimated distribution.  In this study, we consider
a point cloud acquired from a silicone phantom with an inverted
Gaussian surface profile, which is also used later in the real-robot
experiments. This geometry was selected due to its pronounced
curvature and the presence of intrinsic physical boundaries induced by
the surface dimple.  For this simulation the synthetic stiffness field
is defined as
$k(\mathbf{x})=\sum_{i=1}^{2} w_i \,
\frac{\mathcal{N}(\mathbf{x};\boldsymbol{\mu}_i,\boldsymbol{\Sigma}_i)}{\mathcal{N}(\boldsymbol{\mu}_i;\boldsymbol{\mu}_i,\boldsymbol{\Sigma}_i)}
+ k_0$, with $\mathbf{x}\in\mathbb{R}^2$.  Parameters are
$\boldsymbol{\mu}_1=[0.30,\,0.02]^\top$,
$\boldsymbol{\mu}_2=[0.27,\,0.08]^\top$,
$\boldsymbol{\Sigma}_1=\boldsymbol{\Sigma}_2=10^{-4}\mathbf{I}$,
$w_1=40$, $w_2=50$, and $k_0=40$.  This yields two localized
high-stiffness regions embedded in a homogeneous background, visible
in the original distribution~\autoref{fig:alpha_eff}, with
$\alpha=\{0.2,0.4,0.6,0.8\}$, keeping the simulation duration fixed to
$\SI{150}{\second}$. For low $\alpha$, the agent prioritizes
exploration and traverses informative regions without dwelling,
continuing the search for other salient areas. As $\alpha$ increases,
the time spent within relevant regions increases accordingly.
For sufficiently large values of $\alpha$, the search becomes dominated by premature exploitation.
For $\alpha=0.8$ the agent remains near the first informative region it encounters and fails to discover the second one within the fixed simulation horizon. Excessive exploitation can therefore produce incomplete spatial coverage and prevent reconstruction of the complete target field.

\begin{table}[!t]
\centering
\caption{
Effect of $\alpha$ on ergodicity, reconstruction accuracy (RMSE), and convergence time.
Values are reported as mean $\pm$ std over $50$ runs.
Simulations stop when $\mathrm{RMSE} < \SI{2}{k Pa}$ or after $\SI{400}{s}$ if convergence is not achieved.
}
\label{tab:alpha_tradeoff}
\begin{tabular}{c c c c}
\hline
$\alpha$ &
Ergodic Metric &
RMSE [\SI{}{k Pa}] &
Time [s] \\
\hline
0.0 & $0.094 \pm 0.009$ & $2.02 \pm 0.28$ & $344 \pm 42$ \\
0.2 & $0.141 \pm 0.036$ & $1.90 \pm 0.10$ & $136 \pm 55$ \\
0.4 & $0.145 \pm 0.019$ & $1.89 \pm 0.07$ & $108 \pm 16$ \\
0.6 & $0.124 \pm 0.019$ & $1.93 \pm 0.06$ & $182 \pm 27$ \\
0.8 & $0.115 \pm 0.026$ & $3.14 \pm 1.92$ & $384 \pm 21$ \\
\hline
\end{tabular}
\end{table}

We quantified the effect of $\alpha$ on the ergodic metric, accuracy
(RMSE) and convergence time in~\autoref{tab:alpha_tradeoff}. For each
value of $\alpha$, $50$ independent simulations were conducted, and
the mean and standard deviation of each metric were computed.
To compare convergence times under a common accuracy criterion, each simulation was terminated when the reconstruction error fell below $\SI{2}{\kilo\pascal}$, corresponding to approximately $4\,\%$ of the nominal stiffness range, or when the maximum duration of $\SI{400}{\second}$ was reached, whichever occurred first.
The case $\alpha = 0$, corresponding to purely exploratory behavior, i.e., ergodic exploration without local refinement driven by the estimated distribution, is reported as a baseline to provide a direct comparison with the proposed balanced exploration-exploitation strategy. The mean simulation time was found to be less than 400 seconds, with a root mean square error of 2.02. This suggests that stiff areas were not identified in all simulations. Notably, for $\alpha = 0.8$, the $\mathrm{RMSE}$ remains larger than the prescribed threshold, indicating failure to converge to the desired reconstruction accuracy, consistent with persistent exploitation bias and loss of global coverage.

The EID formulation provides a practical guideline for selecting $\alpha$. Since its exploration and exploitation terms are separately normalized, $\alpha$ directly determines their relative contribution. Values close to zero emphasize uncertainty-driven global coverage but may allocate insufficient time to refining informative regions within a finite exploration horizon. Conversely, values close to one suppress uncertainty-driven exploration and can cause the search to concentrate prematurely on the first salient region detected. An intermediate value therefore provides a robust default by preserving both global discovery and local refinement. In the considered scenario, values between $0.2$ and $0.6$ reach the target accuracy, with $\alpha=0.4$ providing the shortest mean convergence time among the tested values.
In general, a practical default is $\alpha=0.4$, which introduces a slight bias toward uncertainty-driven exploration while retaining substantial exploitation of detected informative regions. This conservative choice reflects the asymmetric consequences of parameter selection: an excessively small value primarily slows local refinement, whereas an excessively large value can prevent undiscovered regions from being visited. The parameter can then be increased when exploration is too diffuse or decreased when premature localization is observed.

\subsection{Comparisons with SoA Methods}
We conduct a systematic comparison with representative state-of-the-art
methods to quantify the advantages and limitations of the proposed approach.
In particular, in the next sections (full trajectories and reconstructed maps are available on the paper website) we address the following questions:
\begin{enumerate*}[label=Q\arabic*:]
    \item What is the impact of accounting for system dynamics during exploration?
    \item What is the effect of decoupling exploration from estimation updates?
    \item What is the role of the intrinsic surface geometry and topology?
    \item How does adaptive, information-driven exploration compare with
    prescribed geometric coverage under different travel budgets?
    \item How do diffusion-based and spectral ergodic exploration strategies
    affect convergence and accuracy?
\end{enumerate*}
All comparisons use the plate point cloud with multiple embedded
shapes~(\autoref{fig:gpr_test}c), previously adopted as a benchmark
in~\citet{Bilaloglu2025TactileSurfaces}. Its mild curvature
corresponds to the regime in which BO
baselines~\citep{Garg2016TumorSampling,Ayvali2017Utility-GuidedAbnormalities}
were originally formulated, making the comparison conservative with
respect to the proposed method. The comparative results therefore
isolate the effect of the exploration strategy; indeed, the behavior
on more strongly curved geometries has been shown in
\autoref{fig:shapes_search} and \autoref{fig:alpha_eff}, and the
effect of surface geometry on mapping and planning is deeply assessed
separately in \autoref{sec:kernel_ablation} and
\autoref{sec:uv_comparison}.
The target manifold contains multiple extended and spatially disjoint informative regions, reflecting the objective of reconstructing a complete elasticity distribution rather than locating a single stiffness maximum. This setting is relevant when multiple inclusions or heterogeneous regions may be present, and their spatial extent must be recovered.
The benchmark therefore examines whether each method can discover and reconstruct all informative regions under a matched travel budget, rather than only locate the first or global stiffness maximum. Since both the estimated and reference maps contain normalized, dimensionless values in the range [0, 1], the resulting error metrics, including the RMSE, are also dimensionless.

\begin{figure}[t!]
    \centering
    \includegraphics[width=1.0\linewidth]
    {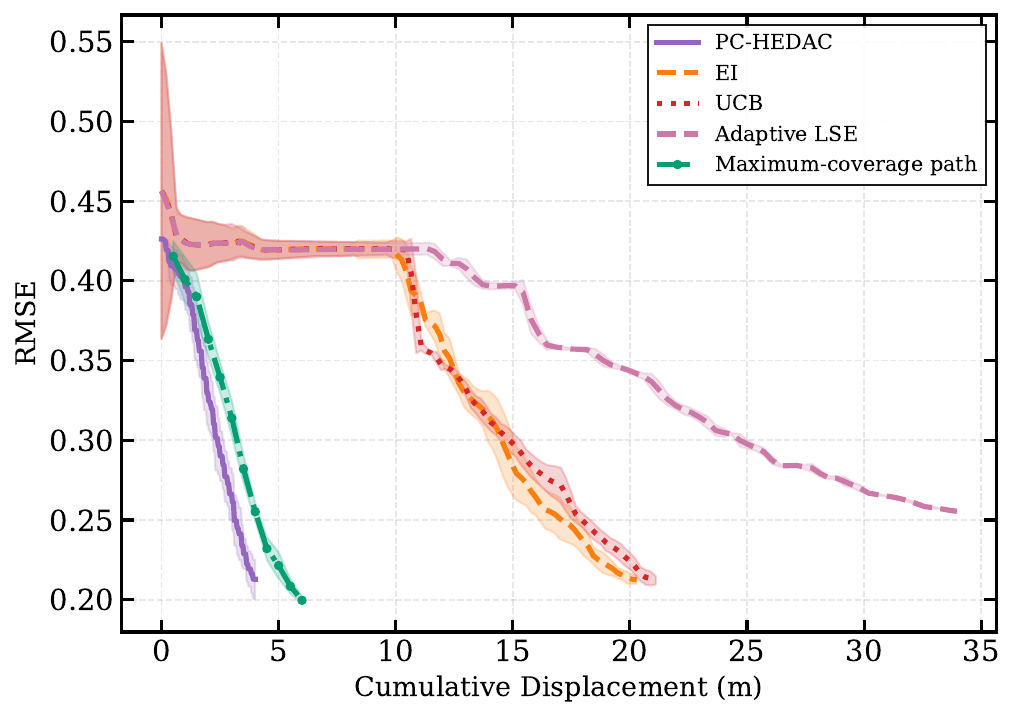}
    \caption{Reconstruction RMSE versus cumulative traveled distance for
    PC-HEDAC, global BO-EI and BO-UCB, adaptive LSE, and the prescribed
    Maximum-Coverage Path (mean $\pm$ standard deviation over ten runs).
    BO and LSE select queries without travel cost in the acquisition; the
    maximum-coverage trajectory is replanned independently for every
    $0.5$~m budget.}
    \label{fig:rmse_evolution}
\end{figure}

\subsubsection{Comparison with Adaptive Acquisition Baselines (Q1, Q2)}
\label{sec:bo_comparison}
Our method incorporates system dynamics during sample generation and
decouples estimation from exploration updates. We evaluate these choices by
comparing our Point Cloud HEDAC (PC-HEDAC) against Bayesian Optimization
with expected improvement
(EI)~\citep{Ayvali2017Utility-GuidedAbnormalities} and
upper confidence bound (UCB)~\citep{Garg2016TumorSampling}, and against
adaptive level-set estimation (LSE)~\citep{Gotovos2013Active}.
All baselines use the same surface-aware GP as our method.

We first consider global variants that select each query by optimizing
the acquisition over all unvisited vertices, without motion cost;
travel is accounted for a posteriori as the geodesic distance between
queries. The acquisition parameters are selected on validation starts
disjoint from the reported runs, via a staged sweep of
$\xi\in\{0.005,0.01,0.025,0.05, 0.1,0.2\}$ for EI and
$\kappa\in\{0.25,0.5,1,2,4,8\}$ for UCB with held-out confirmation,
retaining $\xi=0.005$ and $\kappa=2$.  Over ten trials from shared
initial references, PC-HEDAC reaches an RMSE of $0.2129\pm0.0123$ after $3.997\pm0.001$~m of travel and 40
GP evaluations~(\autoref{fig:rmse_evolution}).  Reaching the same mean
RMSE required $19.77\pm0.55$~m and $274$ GP evaluations for EI, or
$20.68\pm0.32$~m and 259 evaluations for UCB, approximately five times
more travel. Under the same 300-update budget, adaptive LSE plateaus
at $0.2554\pm0.0009$ without attaining PC-HEDAC's accuracy, consistent
with its boundary-classification rather than whole-field
objective. Selecting informative queries without their execution cost
can thus appear sample efficient while demanding substantially more
motion and GP recomputation from a physical agent.

\begin{figure}[t!]
    \centering
    \includegraphics[width=0.98\linewidth]
    {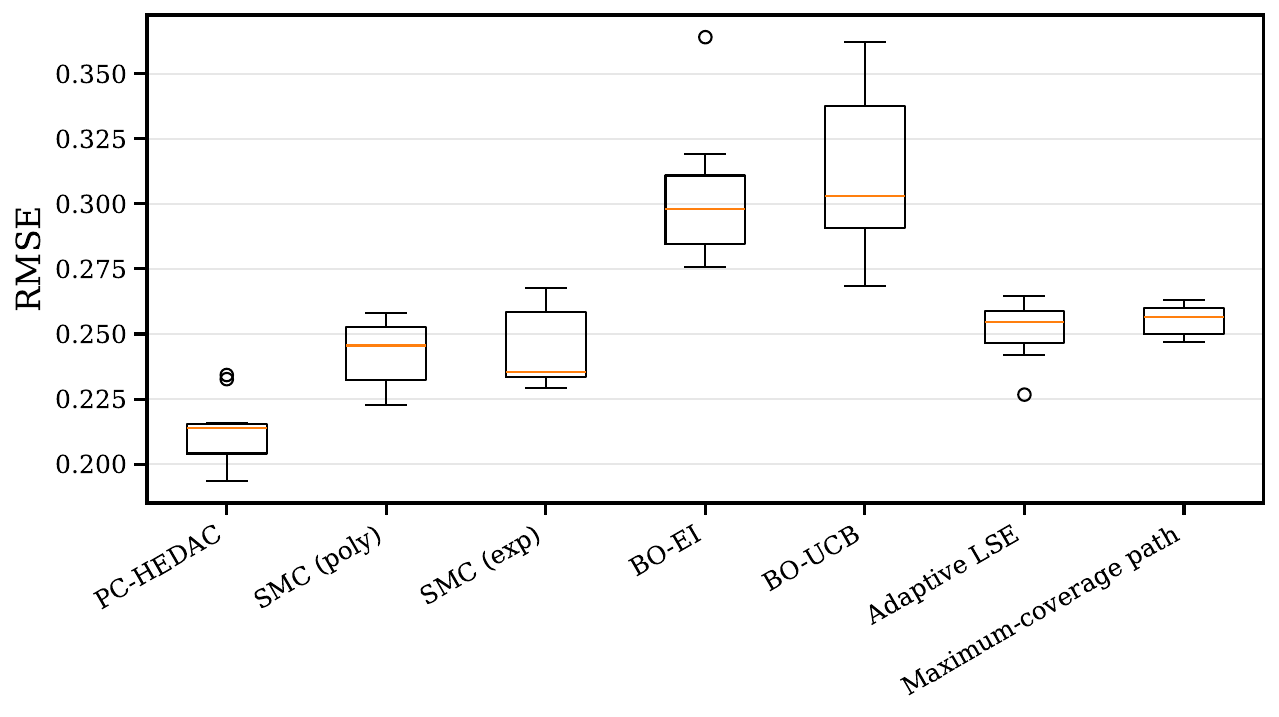}
    \caption{RMSE over ten runs under an approximately $4$~m travel budget.
    PC-HEDAC and the SMC variants use a 2000-step horizon; local BO-EI,
    BO-UCB, and adaptive LSE use continuous surface motion with GP updates
    every 50 steps (LSE reselects ambiguity targets from the cached
    posterior within an interval, without extra GP computations).
    Maximum-Coverage Path uses its independently replanned $4.0$~m
    trajectories.}
    \label{fig:rmse_comparison_ergodic}
\end{figure}

Second, we match PC-HEDAC's budgets with continuous variants sharing its
schedule: maximum displacement $0.002$~m per step, a measurement every five steps, and a GP refit every 50 steps on the cumulative observation set, i.e., 400 measurements and 40 GP updates over 2000 steps, with mean travel
between $3.77$ and $4.00$~m for all methods.
For EI and UCB the acquisition is restricted to candidates within
$0.05$~m, defining the motion direction until the next update; LSE
keeps its global acquisition but executes targets along shortest paths
on the point-cloud graph. Under these matched budgets, PC-HEDAC
achieves the lowest reconstruction RMSE of $0.2129\pm0.0123$, against $0.3022\pm0.0248$ of BO-EI,
$0.3113\pm0.0288$ of BO-UCB, and $0.2514\pm0.0106$ of adaptive
LSE~(\autoref{fig:rmse_comparison_ergodic}). The latter
boundary-targeted sampling clearly outperforms extrema-seeking BO, yet
remains well above PC-HEDAC. The Maximum-Coverage Path and ergodic
baselines (SMC-based) in the figures are analyzed separately in
\autoref{sec:coverage_comparison} and
\autoref{sec:ergodic_comparison}, respectively.

\subsubsection{Comparison on Surface Parameterization (Q3)}
\label{sec:uv_comparison}
\begin{figure*}[t]
  \centering \includegraphics[width=0.95\linewidth]
  {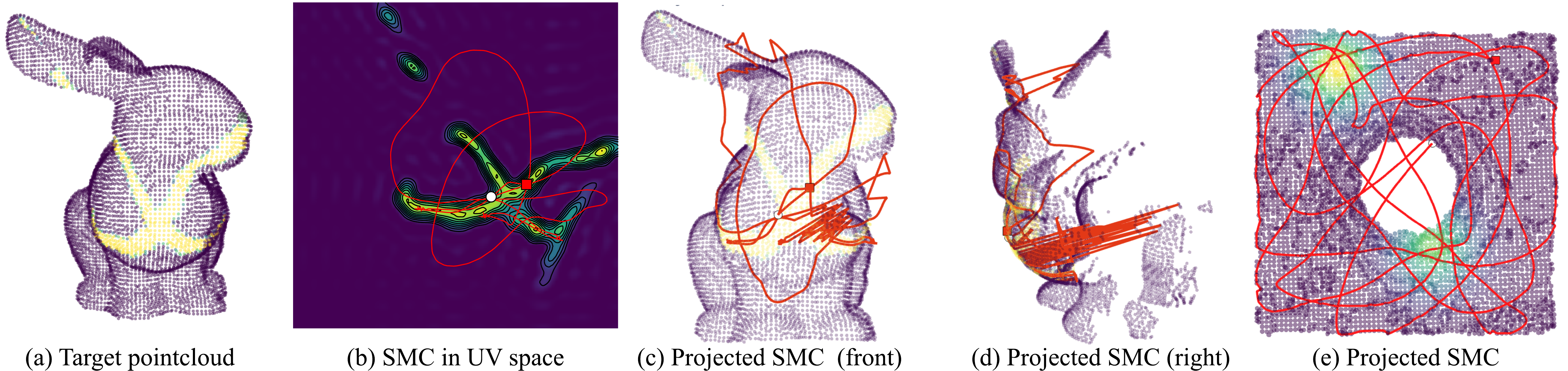}
  \caption{Distortion introduced by UV parameterization in ergodic
    control.  (a) Point cloud with an X-shaped information target
    highlighted in yellow.  (b) Target and exploration trajectory in
    UV space (trajectory in red).  (c,d) Front and right views of the
    SMC trajectory mapped from UV space back onto the surface. (e)
    Failure case on a target object with an internal hole.}
  \label{fig:uv_parameterization}
\end{figure*}

To assess the consequences of planning on a flattened surface
representation, we compare PC-HEDAC with the parameterization-based
ergodic approach of~\citet{Schneyer2026AnPreferences}. The method maps
a watertight surface mesh to a rectangular UV domain and applies SMC
in the resulting parameter space. Because our input is an unmeshed
point cloud, it lacks the face connectivity required to compute the
baseline's UV parameterization. We therefore construct a
point-cloud-compatible baseline by replacing their UV mapping with a
PCA-based UV parameterization and executing SMC in the projected
domain. \autoref{fig:uv_parameterization} exposes two failure modes of
this planar representation.
First, flattening introduces distortion
and tearing, which produce discontinuous trajectories when mapped back
to the surface. Second, the projected domain does not preserve
internal boundaries, allowing the trajectory to cross regions that are
disconnected on the original surface
(\autoref{fig:uv_parameterization}e). PC-HEDAC instead operates
directly on the intrinsic point-cloud geometry, preserving surface
connectivity and both external and internal boundaries. It therefore
generates continuous, surface-constrained trajectories without
requiring a watertight mesh or planar parameterization, as shown
in~\autoref{fig:shapes_search} and~\autoref{fig:alpha_eff}.

\subsubsection{Comparison with Prescribed Coverage Paths (Q4)}
\label{sec:coverage_comparison}
As the surface point cloud and its boundaries are known before exploration, a natural alternative is to precompute a path that maximizes geometric coverage without adapting to the acquired measurements.
We therefore compare PC-HEDAC against a non-adaptive Maximum-Coverage Path defined on the point-cloud graph.
The path follows quantized isocontours of graph-geodesic distance from
a support-border initial position. For each prescribed travel budget,
evaluated every $0.5$~m increment, it is replanned from scratch to
maximize the fraction of vertices covered within the $6$~mm sensing
radius, and measurements follow the same spatial sampling as the
adaptive methods. The mean covered fraction grows from $6.9\%$ at
$0.5$~m to $76.0\%$ at $4.0$~m, where the method obtains an RMSE of
$0.2552\pm0.0057$~(\autoref{fig:rmse_comparison_ergodic}), and it first reaches the target mean
RMSE of $0.2$ at $6.0$~m with $98.1\%$
coverage ($0.1996\pm0.0023$),~\autoref{fig:rmse_evolution}.

These results reveal the limitation of prescribed coverage. At short
travel budgets, the Maximum-Coverage Path provides effective global
exploration and outperforms matched-budget BO while remaining
comparable to adaptive LSE. However, because the path does not adapt
to the acquired measurements, it cannot transition from broad
exploration to the focused refinement of informative regions. It
therefore continues to allocate travel to geometric coverage and
requires $50\,\%$ additional motion, reaching near-complete surface
coverage, before its RMSE falls below $0.2$.
This limitation is particularly relevant to robotic palpation, where additional travel increases both procedure duration and cumulative tissue interaction, potentially increasing patient discomfort. By adapting the trajectory to the estimated information distribution, PC-HEDAC concentrates interaction on informative or uncertain regions and achieves higher reconstruction accuracy with less motion.

\subsubsection{Comparison with Ergodic Planners (Q5)}
\label{sec:ergodic_comparison}
Finally, we compare PC-HEDAC against Laplacian-based SMC planners.
All methods use the same Laplace-Beltrami operator, GP model, EID, and evaluation budget, so the comparison primarily concerns how the coverage error is converted into motion. PC-HEDAC constructs the
pointwise positive visitation deficit $u_0$ and diffuses it using the
full discrete operators. SMC instead projects the signed
target-coverage mismatch onto a truncated Laplacian eigenbasis and
weights the resulting modal errors using either polynomial
$(1+\sqrt{\lambda_k})^{-2}$~\citep{xuMeasurePreservingFlows2024} or exponential
$\exp(-\gamma\sqrt{\lambda_k})$~\citep{dongErgodicExplorationMeshable2025} weighting.

Parameter selection deliberately favors the baselines. The only deviation
from the published SMC configuration, \(\gamma=0.01\) in place of
\(\gamma=0.1\)~\citep{dongErgodicExplorationMeshable2025}, was adopted
because it improves the baseline in our setting. The coefficient
\(\alpha\), which blends the GP posterior mean and standard deviation into the
expected information density in~\eqref{eq:eid},
applies only to the three ergodic controllers and is fixed to
$\alpha=0.2$, identically and without optimization for any of them, so that all ergodic planners cover an identical target.

Under these matched budgets, PC-HEDAC obtains the lowest RMSE
($0.2129\pm0.0123$, followed by
SMC~(poly) ($0.2438\pm0.0144$) and SMC~(exp) ($0.2574\pm0.0244$) as
shown in~\autoref{fig:rmse_comparison_ergodic}.
This gap is consistent with the different representations of the coverage
error: SMC projects it onto a truncated eigenbasis, whereas PC-HEDAC
diffuses it with the full discrete operators,
as described in~\autoref{sec:ergodic_pc}.

Taken together, these comparisons show that PC-HEDAC achieves the
lowest RMSE among all baselines under matched travel and update
budgets. The results support combining continuous, surface-constrained
motion with closed-loop adaptation of the exploration target, rather
than relying on boundary-targeted acquisition, prescribed geometric
coverage, or truncated spectral coverage alone.

\section{Experiments}
While the simulation studies provided a comparative evaluation of the exploration strategies, the physical experiments address a complementary objective: validating the complete closed-loop framework under real contact dynamics, sensing uncertainty, surface-geometry errors, and spatially varying material response. Accordingly, the physical evaluation focuses on execution feasibility, reconstruction repeatability, and stiff-region localization rather than repeating the planner comparison under less controllable conditions.

\subsection{Experimental Setup}
Real-world experiments were carried out using a torque-controlled 7-DoF Franka robotic arm equipped with an Intel RealSense D405 depth camera and BOTA SensONE force–torque sensor mounted at the wrist (see~\autoref{fig:overview}). The software architecture was implemented in ROS~2, with estimation and low-level control modules operating at a control frequency of \SI{1}{k\hertz} and ergodic planner at \SI{500}{\hertz}. Viscoelastic estimates are acquired along the exploration trajectory at
$f_s = \SI{4}{\hertz}$; the GP prediction is executed in parallel every 4 measurements at $\SI{1}{\hertz}$. The PC used is equivalent to the one used in the simulations.
The Cartesian hybrid force-impedance controller introduced in~\autoref{app:hybrid_fi_control} was adopted to regulate the interaction forces during palpation, ensuring compliant physical contact with the tissue phantom (control parameters available in \autoref{app:control_parameters}).
A custom 3D-printed indenter was installed at the tool tip~\citep{Beber2025Force-BasedArm}, featuring a spherical contact interface with a radius of \SI{5}{mm} to guarantee a well-defined contact footprint.

\begin{figure}[t]
    \centering
    \subfloat[Hemispheric phantom.]{
    \includegraphics[trim=0 5mm 0 0, clip, height=2.1cm]{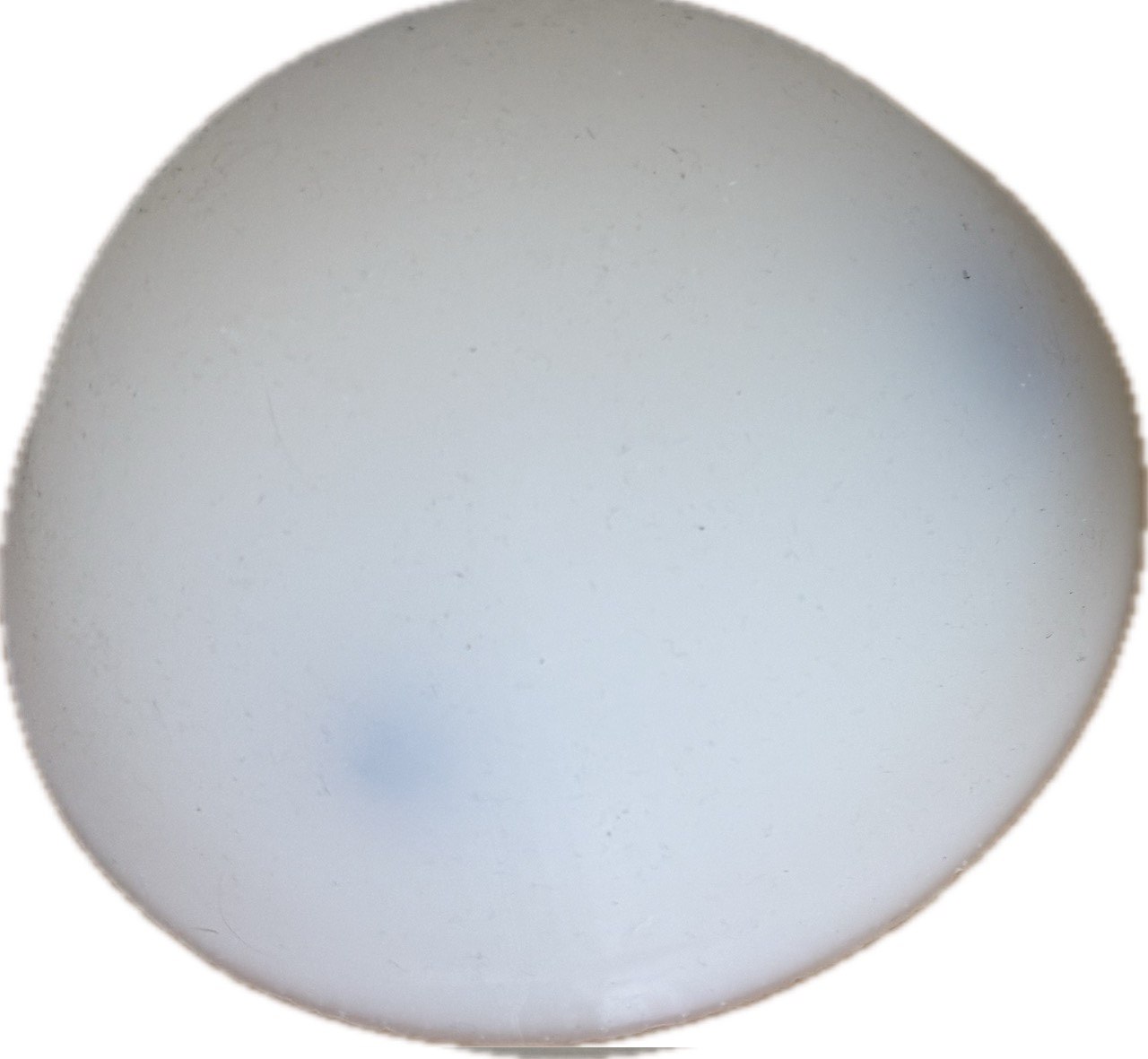}
        \label{fig:hemisphere}
    }
    \hfill
    \subfloat[Gaussian dimple phantom.]{
    \includegraphics[trim=0 5mm 0 0, clip, height=2.1cm]{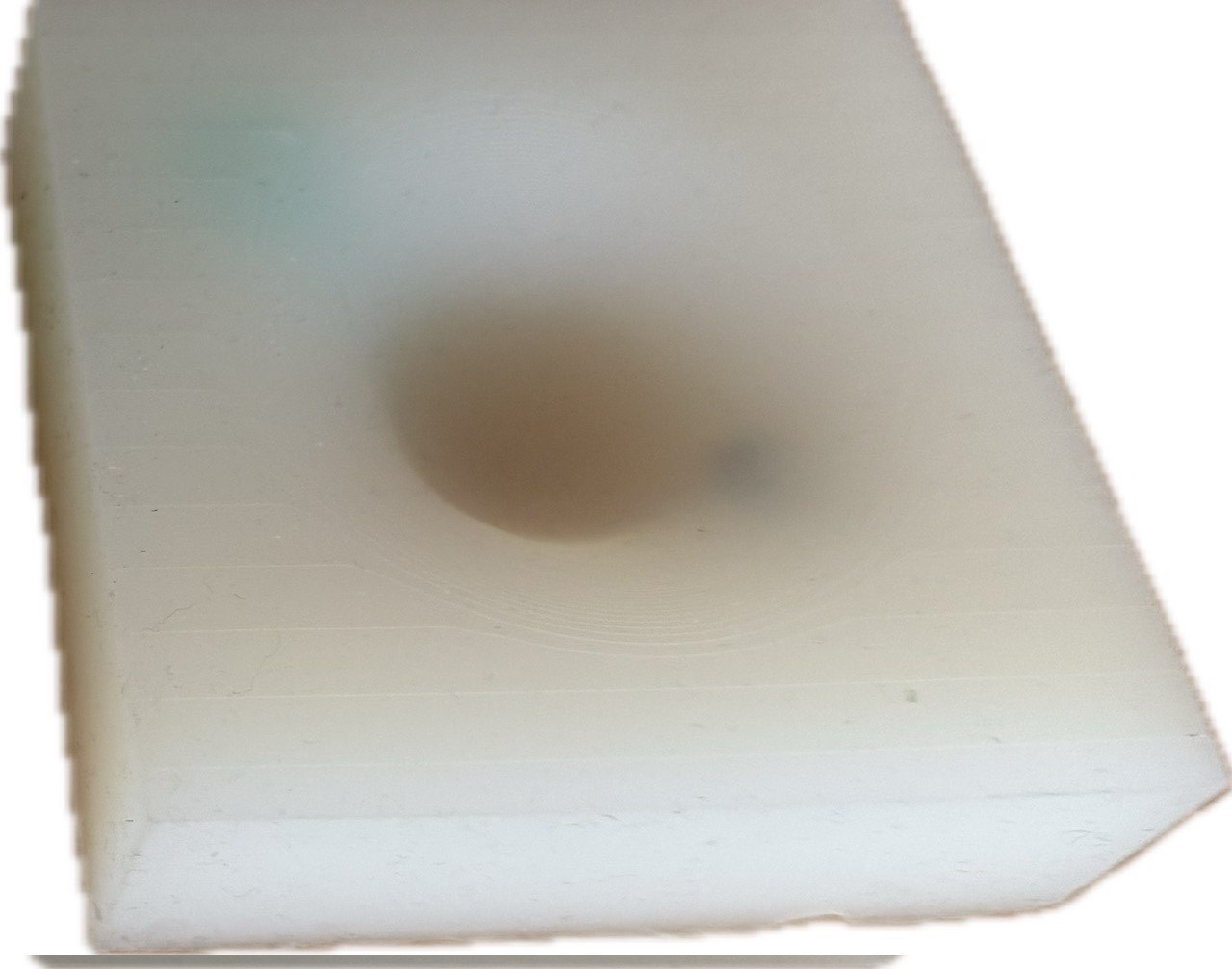}
        \label{fig:dimple}
    }
    \hfill
    \subfloat[Stiffer colored spherical inclusions.]{
\includegraphics[height=2.1cm]{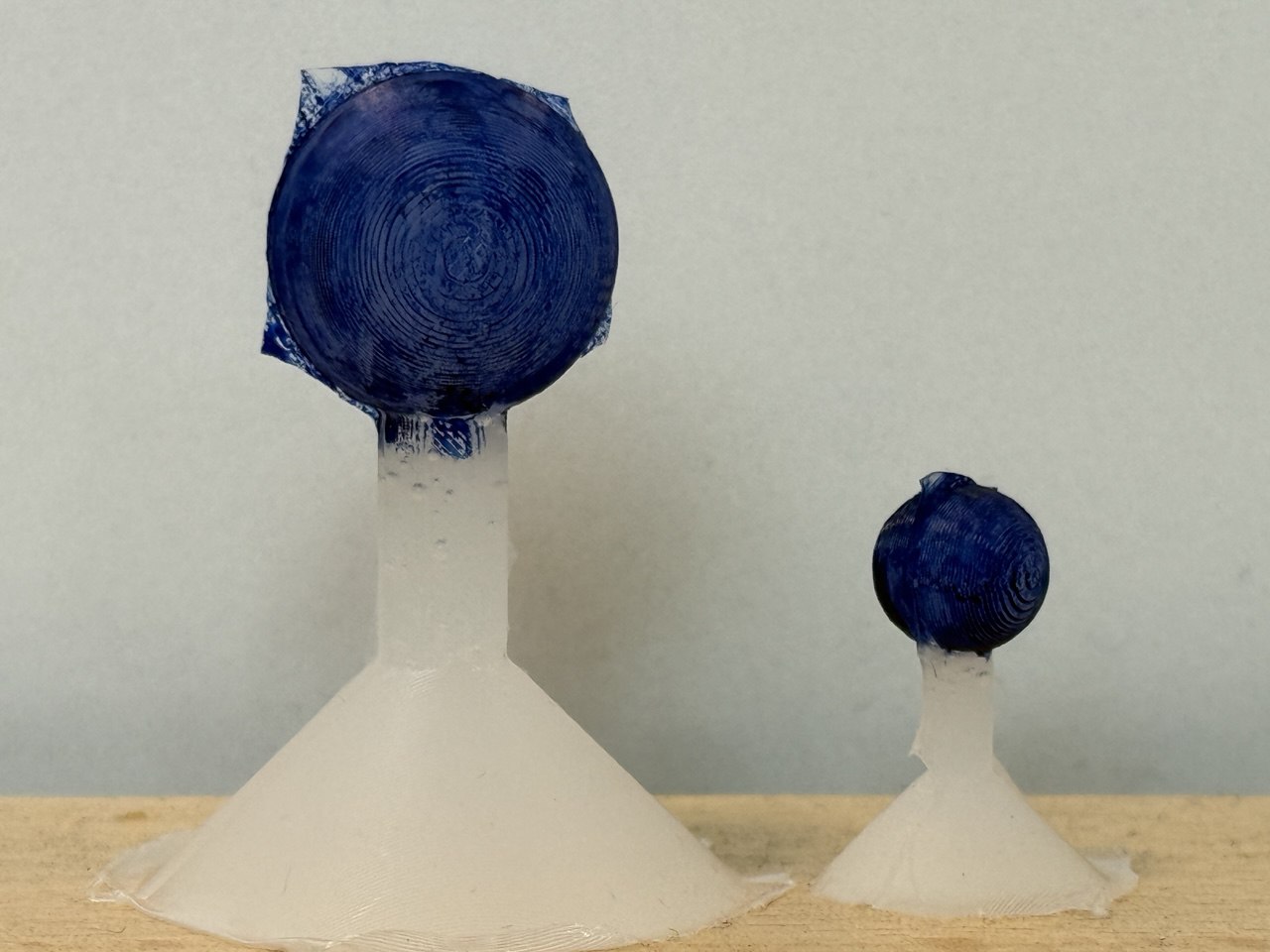}
        \label{fig:spheres}
    }
    \hfill
    \caption{(a-b) Silicone phantoms used for the experiments, (c) stiff inclusion.}
    \label{fig:phantom_fabrication}
\end{figure}

The proposed framework was experimentally evaluated on two
tissue-mimicking silicone samples, a hemispherical phantom and a
Gaussian-dimple phantom (illustrated in~\autoref{fig:phantom_fabrication}),
to test distinct surface geometries and elasticities.
Their stiffness ranges lie within reported values for healthy soft
tissue and stiff lesions~\citep{wells2011medical}. The
\emph{hemispherical phantom} replicates the geometry of a commercial
breast biopsy training phantom.  Its matrix (ECOFLEX-0030 with
\SI{30}{\percent} silicone oil by weight) has an apparent elasticity
of approximately \SI{35}{\kilo\pascal} and embeds three spherical
inclusions (DRAGONSKIN-30NV), two with a radius of
\SI{5}{\milli\meter} and one with a radius of \SI{10}{\milli\meter},
with apparent elasticity between \SIrange{50}{80}{\kilo\pascal}.  The
\emph{Gaussian-dimple phantom} introduces a pronounced concavity and
an internal boundary, testing surface-constrained exploration on a
non-convex domain. Its matrix (ECOFLEX-0030) has an apparent
elasticity of approximately \SI{125}{\kilo\pascal} and embeds a flat
\SI{1.5}{\centi\meter}~$\times$~\SI{1.5}{\centi\meter} inclusion, a
spherical inclusion with a radius of \SI{5}{\milli\meter}, and a
U-shaped lesion of \SI{2}{\milli\meter}~$\times$~\SI{5}{\milli\meter},
with apparent elasticity between \SIrange{140}{190}{\kilo\pascal}.
Their fabrication process is detailed in~\autoref{sec:phantom} to
support reproducibility.

The GP lengthscale was fixed to $l=\SI{5}{mm}$, corresponding to the
radius $R=\SI{5}{mm}$ of the spherical indenter used in the
experiments (full parameters in the~\autoref{app:control_parameters}) and the
number of the Laplace-Beltrami eigenpairs to $n_{\mathrm{eig}}=500$.
Since the measured force arises from distributed deformation beneath
the indenter rather than from an infinitesimal point contact, each
local viscoelastic estimate reflects the mechanical response averaged
over a finite contact region. The indenter therefore defines the
effective spatial resolution of the palpation measurement, acting as a
mechanical low-pass filter on the stiffness information acquired from
the surface. Choosing the GP lengthscale at this scale provides a
physically motivated smoothness prior and avoids introducing spatial
variations finer than those supported by the contact measurement
itself.

The point cloud is acquired once, before exploration, from a single
depth image taken with the camera positioned approximately
$10\,\mathrm{cm}$ above the center of the sample and oriented along
its mean surface normal. Points not belonging to the silicone sample
are removed, and the retained points define the admissible domain from
which the Laplace-Beltrami decomposition and the kernel lookup table
are computed. The cloud is not updated thereafter. The camera is
rigidly mounted to the end effector through a 3D-printed bracket, so
the camera--robot transformation is fixed and it is obtained directly
from the CAD models of the camera and the mounting point, declared in
the robot URDF.  The initial contact is established by manually
guiding the arm until the indenter rests on the surface, after which
the exploration runs autonomously. This mirrors the intended clinical
workflow, in which an operator positions the probe on the region to be
examined, and the system performs the examination. The exploration
begins with the indenter already in contact with the surface, at an
initial position that differs among the trials, and the contact is
maintained continuously throughout the prescribed exploration horizon,
while data are acquired without interruption.

\subsection{Evaluation and Results}
\label{sec:experimental_reference}
The physical experiments evaluate three aspects of the proposed
framework: the repeatability of the reconstructed elasticity maps from
different initial contact positions, the accuracy with which the
embedded stiff regions are reconstructed and localized, and the
time-accuracy trade-off introduced by the stopping criterion.
Reconstruction accuracy is evaluated against independently acquired
pointwise reference maps using whole-map RMSE and region-wise
localization error. Eight palpation trials were conducted on each phantom from different initial contact positions and continued for fixed exploration budgets of $\SI{500}{\second}$ for the hemispherical phantom and $\SI{300}{\second}$ for the Gaussian-dimple phantom.
These budgets correspond to the estimated time required to cover the
accessible surface continuously at the same tangential speed and
spatial sampling resolution used by the proposed method, without the
dwell times and retraction motions required by pointwise probing
(required in the computation of the elasticity reference map).
Results are evaluated at two time instants: at the stopping time
$T_{\mathrm{STOP}}$, detected online by the designed proposed
criterion, and at the end of the full exploration budget
$T_{\mathrm{FULL}}$.

\subsubsection{Physical Reference Map}
Unlike the simulations, where the elasticity distribution is
prescribed and known exactly, the elasticity distribution of a
fabricated phantom is not directly available and must be characterized
experimentally.
The pointwise DRM+UKF procedure used to construct the reference maps
was previously validated against standard mechanical testing on
homogeneous silicone specimens (the filter parameters are reported in
the~\autoref{app:control_parameters}), showing agreement with an
Instron 4502 universal testing machine-derived elasticity estimates
under comparable contact
conditions~\citep{Beber2025Force-BasedArm}. Consequently, a uniformly
sampled pointwise map provides a reliable experimental reference for
the mechanical response estimated by the adopted sensing modality. The
present comparison specifically quantifies the additional error
introduced by the continuous exploration and the spatial
reconstruction, while it does not constitute a new validation of the
absolute material-property estimates.

Since the pointwise measurements are available only at the grid
points, a separate heat-kernel GP, with the same length scale as the
online mapping procedure and the prior mean set to the average of the
measurements outside the inclusion regions, interpolates them over the
processed surface. Its posterior mean~\eqref{eq:gp_mean} defines the empirical reference
map $\bs{\mu}_{\mathrm{ref}}$.

\subsubsection{Stopping Criterion and Time--Accuracy Analysis}
\label{sec:stopping_validation}

To determine if the exploration can be stopped, we monitor whether the accumulated visitation distribution has
converged to the current EID. A small discrepancy indicates that the
accumulated visitation is close to the distribution currently prescribed
by the evolving information objective, whereas a stable discrepancy
indicates that continued exploration is producing limited changes in
coverage. This criterion is used as an online proxy for convergence and
is subsequently assessed against reconstruction and localization
accuracy.

The ergodic metric~\eqref{eq:coverage_metric} is used only for
termination and does not enter the exploration controller. A candidate
stopping time is detected when two conditions are satisfied. First,
$em(t)<0.3$, corresponding to a reduction of at least $70\,\%$
relative to its normalized initial value. Second, the mean rate of
change of $em(t)$ over the preceding $\SI{20}{\second}$ being above
$-10^{-3}$, indicating that the discrepancy is no longer decreasing
appreciably while suppressing short-term fluctuations. For evaluation
purposes, this criterion was monitored online without interrupting the
experiment, such that each trial provides both a reconstruction at the
candidate stopping time and one at the end of the prescribed
exploration budget.  The criterion triggered after
$367.91 \pm 24.20\,$\si{s} on the hemispherical phantom and
$195.71 \pm 35.91\,$\si{s} on the Gaussian-dimple phantom,
corresponding to reductions in exploration time of
$26.42 \pm 4.84\,$\si{\percent} and $34.76 \pm 11.97\,$\si{\percent},
respectively, relative to the fixed exploration budgets.

\begin{table*}[t]
\centering
\caption{Effect of the ergodic stopping criterion on exploration time,
whole-map elasticity RMSE, and per-region localization error.
$T_{\mathrm{STOP}}$ results correspond to the maps at the stopping time
detected by the proposed criterion, whereas $T_{\mathrm{FULL}}$ results
correspond to the end of the fixed exploration budget of
\SI{500}{\second} for the hemispherical phantom and \SI{300}{\second} for the
Gaussian-dimple phantom. Localization error is the three-dimensional
Euclidean distance between the reconstructed elasticity-weighted centroid
and the corresponding stiffness-peak position identified on the
stationary pointwise reference map. Values are reported as mean
$\pm$ standard deviation across eight trials per phantom.}
\label{tab:full_vs_stopped_errors}
\small
\setlength{\tabcolsep}{5pt}

\begin{tabular}{lccccclcc}
\toprule
Sample &
\multicolumn{3}{c}{Exploration time} &
\multicolumn{2}{c}{Elasticity RMSE [kPa]} &
Region &
\multicolumn{2}{c}{Localization error [mm]} \\
\cmidrule(lr){2-4}
\cmidrule(lr){5-6}
\cmidrule(lr){8-9}
& $T_{\mathrm{FULL}}$ [s] & $T_{\mathrm{STOP}}$ [s] & Reduction [\%] &
$T_{\mathrm{FULL}}$ & $T_{\mathrm{STOP}}$ &
& $T_{\mathrm{FULL}}$ & $T_{\mathrm{STOP}}$ \\
\midrule

\multirow{3}{*}{Hem.}
& \multirow{3}{*}{$500$}
& \multirow{3}{*}{$367.91 \pm 24.20$}
& \multirow{3}{*}{$26.42 \pm 4.84$}
& \multirow{3}{*}{$7.654 \pm 0.407$}
& \multirow{3}{*}{$7.602 \pm 0.542$}
& Region 1 & $2.46 \pm 1.52$ & $1.93 \pm 0.75$ \\
& & & & & &
Region 2 & $3.87 \pm 1.31$ & $5.21 \pm 2.51$ \\
& & & & & &
Region 3 & $3.35 \pm 1.15$ & $3.31 \pm 0.85$ \\

\midrule

\multirow{3}{*}{G. D.}
& \multirow{3}{*}{$300$}
& \multirow{3}{*}{$195.71 \pm 35.91$}
& \multirow{3}{*}{$34.76 \pm 11.97$}
& \multirow{3}{*}{$16.686 \pm 1.577$}
& \multirow{3}{*}{$16.059 \pm 1.123$}
& Region 1 & $1.94 \pm 0.93$ & $2.42 \pm 1.59$ \\
& & & & & &
Region 2 & $2.22 \pm 1.01$ & $3.35 \pm 2.88$ \\
& & & & & &
Region 3 & $5.01 \pm 1.35$ & $5.39 \pm 1.42$ \\

\bottomrule
\end{tabular}
\end{table*}

\subsubsection{Reconstruction Error at the Stopping Time}

At the stopping time, the reconstruction error of experiment $k$ is
defined as
\begin{equation}
    \mathrm{RMSE}_k
    =
    \sqrt{
    \frac{1}{n_{\mathcal{P}}}
    \left\|
    \bs{\mu}_k - \bs{\mu}_{\mathrm{ref}}
    \right\|_2^2
    },
    \label{eq:physical_gt_rmse}
\end{equation}
where $n_{\mathcal{P}}$ is the number of vertices in the processed point
cloud. The elasticity RMSE at the stopping time was
$7.602 \pm 0.542\,$\si{kPa} for the hemispherical phantom and
$16.059 \pm 1.123\,$\si{kPa} for the Gaussian-dimple phantom. Although
the absolute RMSE was larger for the Gaussian-dimple phantom, the two
phantoms have different stiffness ranges and contrasts, and their
absolute RMSE values are therefore not directly comparable. More
importantly, the standard deviation was small relative to the mean error
for both phantoms, indicating repeatable reconstruction across trials
initialized from different contact positions.

To assess whether the additional exploration beyond the stopping time
produced a consistent improvement in reconstruction accuracy, the
stopping-time and full-budget reconstructions were compared on a
trial-by-trial basis. The stopping parameters were selected after
inspection of these trials; therefore, this analysis is exploratory
rather than a prospective validation.
For each trial $k$, the paired RMSE difference was defined as
\begin{equation}
    \Delta \mathrm{RMSE}_k
    =
    \mathrm{RMSE}_{\mathrm{stop},k}
    -
    \mathrm{RMSE}_{\mathrm{full},k},
    \label{eq:paired_rmse_difference}
\end{equation}
such that positive values indicate lower reconstruction error at the
full exploration horizon. The $95\,\%$ confidence interval (CI) of the
mean paired difference was computed using a two-sided Student's
$t$-interval,
\begin{equation}
    \overline{\Delta\mathrm{RMSE}}
    \pm
    t_{0.975,n-1}
    \frac{s_{\Delta}}{\sqrt{n}},
    \label{eq:paired_rmse_ci}
\end{equation}
where $\overline{\Delta\mathrm{RMSE}}$ and $s_{\Delta}$ denote,
respectively, the sample mean and sample standard deviation of the
paired differences across the $n$ trials.  The mean paired differences
were $-0.052\,$\si{kPa} (95\% CI: $[-0.402,\,0.297]\,$\si{kPa}; $n=8$)
for the hemispherical phantom and $-0.627\,$\si{kPa} (95\% CI:
$[-1.603,\,0.349]\,$\si{kPa}; $n=8$) for the Gaussian-dimple
phantom. Thus, continued exploration did not produce a consistent
reduction in the whole-map RMSE.  Similarly, continuing the
exploration to the full horizon changed the mean localization errors
between $-0.53$ and \SI{1.34}{mm} across regions. These changes were
region-dependent and remained below the effective \SI{5}{mm} spatial
resolution of the palpation measurements, suggesting only a marginal
localization benefit from the additional exploration.

Although this analysis cannot establish that early stopping preserves
equivalent reconstruction accuracy, it provides exploratory evidence
that the proposed criterion identifies an earlier operating point with
only small mean changes in reconstruction and localization error relative
to the full prescribed exploration horizon.

\begin{figure*}[t]
    \centering
    \begin{tikzpicture}
        \node (img) {
            \begin{minipage}{\linewidth}
                \centering
                \includegraphics[scale=0.88]{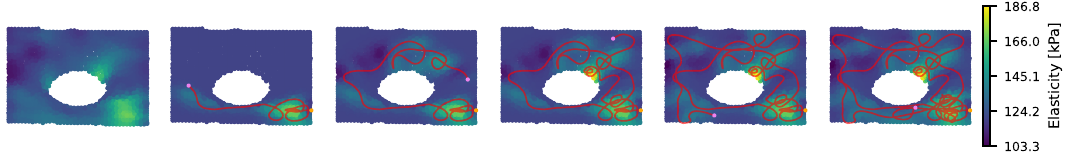}\\
                \includegraphics[scale=0.88]{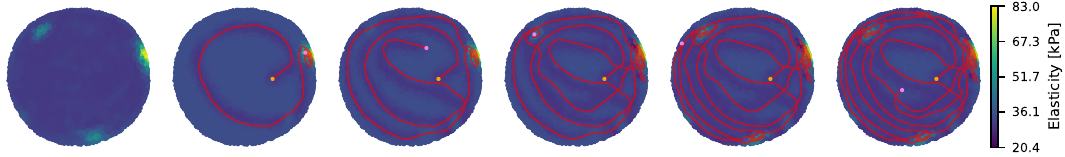}\\
            \end{minipage}
        };

        \node[rotate=90, anchor=east] at ($(img.north west)+(0.5cm,-0.4cm)$) {\footnotesize Gaussian Dimple};
        \node[rotate=90, anchor=east] at ($(img.north west)+(0.5cm,-3.0cm)$) {\footnotesize Hemisphere};

        \node[anchor=west] at ($(img.north west)+(0.2cm,-0.3cm)$) {\footnotesize \textbf{Original Distribution}};
        \node[anchor=west] at ($(img.north)+(0cm,-0.3cm)$) {\footnotesize \textbf{Time Evolution}};

        \draw[red, thick] ($(img.south west)+(1.5cm,2.0cm)$) circle (0.2cm);
        \node[red, anchor=west] at ($(img.south west)+(1.8cm,2.0cm)$) {};

        \draw[red, thick] ($(img.south west)+(2.2cm,0.4cm)$) circle (0.2cm);
        \node[red, anchor=east] at ($(img.south west)+(1.9cm,0.4cm)$) {};

        \draw[red, thick] ($(img.south west)+(3.1cm,1.6cm)$) circle (0.35cm);
        \node[red, anchor=west] at ($(img.south west)+(3.5cm,1.6cm)$) {};

        \draw[red, thick] ($(img.south west)+(1.5cm,3.7cm)$) circle (0.15cm);
        \node[red, anchor=west] at ($(img.south west)+(0.5cm,3.3cm)$) {};

        \draw[red, thick] ($(img.south west)+(2.4cm,3.9cm)$) circle (0.2cm);
        \node[red, anchor=west] at ($(img.south west)+(2.5cm,3.9cm)$) {};

        \draw[red, thick] ($(img.south west)+(2.8cm,3.2cm)$) circle (0.35cm);
        \node[red, anchor=west] at ($(img.south west)+(3.1cm,2.9cm)$) {};

    \end{tikzpicture}

    \caption{Temporal evolution of the reconstructed elastic map
      during ergodic palpation of the Gaussian dimple and hemisphere
      phantoms. From left to right: reference distribution,
      progressive reconstruction, and exploration trajectory. The
      method accurately localizes stiff regions and converges to a
      stable estimate, consistent with the quantitative results
      in~\autoref{tab:full_vs_stopped_errors}. Example explorations
      and online map reconstructions are also available in the
      multimedia attachment.}
    \label{fig:experiments}
\end{figure*}

\subsubsection{Localization Accuracy}
Accurate localization is important because stiff regions must provide
reliable targets for subsequent diagnostic measurements or focused
palpation. Region-wise localization is evaluated using the same
empirical reference map $\bs{\mu}_{\mathrm{ref}}$, obtained by
stationary pointwise palpation, that is used for the whole-map RMSE
evaluation.

On each pointwise reference map, three spatially separated stiffness
maxima are identified by non-maximum suppression with a minimum
separation of \SI{15}{mm}, corresponding to three indenter radii.
Let $\bs{c}_{r,\mathrm{ref}}$ denote the three-dimensional position of
reference maximum $r$, expressed in the same coordinate frame as the
reconstructed surface. Each maximum defines a Euclidean neighborhood
of radius \SI{11.25}{mm}. Vertices belonging to overlapping neighborhoods
are assigned to the nearest reference maximum, yielding the evaluation
regions $\mathcal{R}_r$. The reference positions and evaluation regions
remain fixed across all trials and are used for both stopping-time and
full-budget reconstructions.

For each region $r$ and experimental map $k$, the local background
stiffness is estimated by the 10th percentile of the reconstructed
elasticity,
\begin{equation}
    b_{r,k}
    =
    Q_{0.1}\!\left(
    \left\{\mu_{k,i}\mid i\in\mathcal{R}_r\right\}
    \right),
\end{equation}
where $Q_{0.1}$ denotes the empirical quantile of order $0.1$.
The vertices used to estimate the reconstructed center are those
whose stiffness reaches at least half the local peak prominence,
\begin{equation}
    \mathcal{M}_{r,k}
    =
    \left\{
    i\in\mathcal{R}_r
    \;\middle|\;
    \mu_{k,i}
    \geq
    b_{r,k}
    +
    0.5\left(\mu_{r,k}^{\max}-b_{r,k}\right)
    \right\},
\end{equation}
where
$\mu_{r,k}^{\max}=\max_{i\in\mathcal{R}_r}\mu_{k,i}$.

The reconstructed center is defined as the centroid weighted by
stiffness above the estimated local background,
\begin{equation}
\bs{c}_{r,k}
=
\frac{
\displaystyle\sum_{i \in \mathcal{M}_{r,k}}
\left(\mu_{k,i}-b_{r,k}\right)\bs{p}_i
}{
\displaystyle\sum_{i \in \mathcal{M}_{r,k}}
\left(\mu_{k,i}-b_{r,k}\right)
},
\label{eq:elasticity_weighted_centroid}
\end{equation}
where $\bs{p}_i$ is the three-dimensional position of vertex $i$.
The localization error is the Euclidean distance between this
reconstructed centroid and the corresponding stiffness-peak position
on the pointwise reference map,
\begin{equation}
d_{r,k}
=
\left\|
\bs{c}_{r,k}-\bs{c}_{r,\mathrm{ref}}
\right\|_2 .
\label{eq:physical_centroid_error}
\end{equation}

At the stopping time, the mean localization errors of the individual
regions range from \SI{1.93}{mm} to \SI{5.39}{mm}
(\autoref{tab:full_vs_stopped_errors}).  On the hemispherical phantom,
the errors are $1.93 \pm 0.75\,$\si{mm}, $5.21 \pm 2.51\,$\si{mm}, and
$3.31 \pm 0.85\,$\si{mm} for Regions~1--3, respectively.  On the
Gaussian-dimple phantom, the corresponding errors are
$2.42 \pm 1.59\,$\si{mm}, $3.35 \pm 2.88\,$\si{mm}, and
$5.39 \pm 1.42\,$\si{mm}.

These results show that the position of the stiff regions can remain
localized within a few millimeters even when the reconstructed
elasticity amplitude differs from the physical reference map. In
particular, the phantom with the larger whole-map RMSE does not exhibit
a proportional degradation in localization accuracy, indicating that
spatial localization is less sensitive than absolute amplitude
reconstruction to residual mapping errors. \autoref{fig:experiments} shows the temporal evolution of two representative
experiments, illustrating how the exploration progresses and the map is
reconstructed over time.

\section{Discussion and Conclusion}
\label{sec:conclusion}

This work presented a closed-loop framework for robotic palpation and
elastic mapping directly on curved point-cloud surfaces. Its main
methodological contribution is the use of the same intrinsic geometric
representation for probabilistic mapping and continuous exploration,
thus avoiding global surface parameterization and allowing spatial
correlations, boundaries, and robot motion to be treated consistently
on non-planar domains.

Under matched travel and GP-update budgets, PC-HEDAC achieved the
lowest reconstruction error among all evaluated methods, outperforming
the state-of-the-art BO, adaptive LSE, and SMC baselines, as well as
prescribed maximum coverage. Moreover, the global BO variants required
approximately five times more travel to reach the same mean
reconstruction accuracy. The comparisons match the modeling components
and budgets not under evaluation, so the observed performance
differences primarily reflect how each method translates the available
information into motion. The results therefore support the proposed
combination of continuous surface-constrained exploration,
full-operator diffusion, and closed-loop adaptation of the exploration
target.

Beyond this overall performance advantage, the results provide two
main insights into the formulation. First, the value of the intrinsic
kernel depends on the geometry: its additional construction cost is
justified when ambient Euclidean proximity does not represent surface
connectivity, whereas its advantage is limited on compact, weakly
curved surfaces. Second, selecting informative samples alone is not
sufficient for efficient robotic palpation. Prescribed coverage
supports initial global exploration but cannot subsequently
concentrate interaction on regions revealed by the measurements, while
query-based and spectrally truncated strategies incur greater motion
or lose fine spatial structure. The parameter studies also provide
practical design guidance. Frequent GP updates are fundamental to
prevent the planner from following an outdated information
distribution. An intermediate exploration coefficient provides a
suitable initial balance between global discovery and local
refinement: excessive exploitation can leave informative regions
undiscovered, whereas excessive exploration mainly delays their
refinement. The value identified in the simulations should therefore
be regarded as an initial setting to refine for the considered
geometry and available exploration horizon, rather than as a universal
optimum.

The physical experiments demonstrate that the complete
estimator-mapper-planner loop can repeatedly reconstruct and localize
stiff regions from different initial contact positions. Importantly,
localization remains accurate even when residual amplitude errors are
present in the reconstructed map. This distinction is relevant to
targeted palpation, where the principal output may be a reliable
location for subsequent focused examination rather than an absolute
estimate of tissue elasticity. The candidate stopping criterion also
identifies an earlier operating point with only small mean changes in
reconstruction and localization relative to the prescribed exploration
horizon. Because its thresholds were selected from the same trials,
however, this result is exploratory and requires prospective validation on independent experiments.

The proposed framework nevertheless relies on few modeling assumptions and methodological limitations that delimit its current operating regime.
First, the precomputed intrinsic representation assumes that surface geometry remains approximately unchanged during exploration.
This assumption is compatible with short, targeted examinations of accessible anatomical regions, such as breast or thyroid palpation, where stable patient positioning, regulated contact forces, and instructions to remain still can limit changes in surface geometry during a single exploration. When substantial stretching, folding, respiratory or swallowing motion, or displacement of the examined region cannot be neglected, online surface tracking and adaptation of the geometric model would be required.
A second limitation concerns the interpretation and validation of the
reconstructed mechanical response. Because measurements are acquired
through a finite-size indenter, they represent a spatially averaged,
force-conditioned apparent response rather than a direct constitutive
material property. The underlying force-based estimator was previously
validated against standard mechanical testing on homogeneous specimens~\citep{Beber2025Force-BasedArm}, supporting its use for pointwise
measurements under controlled normal contact. Nevertheless, the
pointwise reference and online reconstruction share the same sensing
hardware, contact model, and estimator. Systematic errors arising from
these common components may therefore affect both maps similarly and
remain undetected. Their comparison consequently quantifies the
additional error introduced by continuous online exploration, rather
than the absolute accuracy of the reconstructed mechanical properties.

Finally, the method maps mechanical response only over the observed
surface and does not recover the depth, three-dimensional geometry, or
clinical significance of subsurface structures. It should therefore be
regarded as an adaptive surface-sensing and localization component
rather than a standalone diagnostic system. Future work will address
online geometric adaptation, prospective validation of the stopping
criterion, and evaluation on more realistic anatomical models.
Integration with robotic ultrasound is a natural extension for using
the identified surface regions to guide depth-resolved imaging~\citep{jiang2023robotic,sigrist2017ultrasound,
jiang2021deformation}.
Within this operating regime, the results demonstrate that intrinsic,
closed-loop ergodic exploration can provide efficient and repeatable
mechanical mapping of curved surfaces under controlled contact.

\begin{acks}
The authors acknowledge the use of ChatGPT (OpenAI) in developing parts of
the code and the text of this manuscript. The authors reviewed and edited
all AI-assisted content and take full responsibility for the final work.
\end{acks}

\begin{dci}
The authors declared no potential conflicts of interest with respect to
the research, authorship and/or publication of this article.
\end{dci}

\begin{sm}
    Code and additional materials will be released upon acceptance. A
    multimedia extension showing example explorations and online map
    reconstructions accompanies this article.
\end{sm}

\bibliographystyle{SageH}
\bibliography{references/references,references/manual_ref,references/cems_refs}

\appendix

\section{Hybrid Force-Impedance Control}\label{app:hybrid_fi_control}

To maintain a constant normal contact force while ensuring compliant tangential motion, a hybrid force-impedance controller is employed, with force control along the surface normal and impedance control along the complementary directions. Although formulations preserving passivity~\citep{Schindlbeck2015UnifiedTanks} or improving performance~\citep{iskandar2023hybrid} have been proposed, a standard implementation~\citep{Villani2016ForceControl} is adopted here.

The task space is decomposed into force- and impedance-controlled
subspaces through projection matrices $\bs{P}_f$ and
$\bs{P}_i \in \mathrm{R}^{6\times 6}$, defined in the end-effector
(EE) frame, such that
\begin{equation}
\bs{P}_f + \bs{P}_i = \bs{I}, \qquad \bs{P}_f \bs{P}_i = \bs{0}.
\end{equation}
Let ${}^{\mathcal{W}}\bs{R}_{\mathcal{EE}} \in \mathrm{SO}(3)$ denote the rotation matrix from the EE frame to the world frame. The corresponding rotation for Cartesian wrenches is
\begin{equation}
\bs{R}_6 = \operatorname{diag}\left({}^{\mathcal{W}}\bs{R}_{\mathcal{EE}},{}^{\mathcal{W}}\bs{R}_{\mathcal{EE}}\right) \in \mathrm{R}^{6x6}.
\end{equation}

\subsection{Cartesian Force Controller}\label{app:cart_force_control}

Force control directly regulates the interaction wrench in the
force-controlled subspace:
\begin{equation}
\bs{F}_f(t) = \bs{P}_f \left[
    \bs{F}_d(t)
    + \bs{K}_p \bs{e}_F(t)
    + \bs{K}_i \int_0^t \bs{e}_F(\tau)\,\mathrm{d}\tau
\right],
\end{equation}
where $\bs{e}_F(t) = \bs{F}_d(t) - \bs{F}_{\mathrm{meas}}(t)$
is the wrench tracking error, and $\bs{F}_d(t)$ and
$\bs{F}_{\mathrm{meas}}(t)$ are the desired and measured EE
wrenches, respectively. The diagonal matrices
$\bs{K}_p,\bs{K}_i \in \mathbb{R}^{6\times6}$ contain the
proportional and integral gains, respectively. The integral
state is initialized to zero when contact control begins. The corresponding joint-space torque $\bs{\tau}_f \in \mathrm{R}^{7}$ is
\begin{equation}
\bs{\tau}_f = \bs{J}^\top(\bs{q}) \bs{R}_6 \bs{F}_f.
\label{eq:force}
\end{equation}
Here, $\bs{J}(\bs{q})$ denotes the geometric Jacobian.

\subsection{Cartesian Impedance Controller}\label{app:cart_imp_control}

In the impedance-controlled subspace, the Cartesian impedance law is
\begin{equation}
\bs{\tau}_i = \bs{J}^\top(\bs{q}) \left(\bs{K}_d \tilde{\bs{x}} + \bs{D}_d \tilde{\dot{\bs{x}}}\right),
\label{eq:impedance}
\end{equation}
where $\tilde{\bs{x}}$ is the Cartesian pose error from the current to the desired EE pose and $\tilde{\dot{\bs{x}}} = \dot{\bs{x}}_d - \dot{\bs{x}}_{EE}$ is the corresponding Cartesian velocity error. The stiffness and damping matrices are specified in the EE frame and projected onto the impedance-controlled subspace:
\begin{align}
\bs{K}_d &= \bs{R}_6 \bs{P}_i \bs{K}_d^{\mathcal{EE}} \bs{P}_i^T \bs{R}_6^T, \\
\bs{D}_d &= \bs{R}_6 \bs{P}_i \bs{D}_d^{\mathcal{EE}} \bs{P}_i^T \bs{R}_6^T.
\end{align}
The damping matrix $\bs{D}_d^{\mathcal{EE}}$ is computed using the double-diagonalization method of~\citep{Albu-Schaffer2003CartesianDLR-Light-Weight-Arms}, ensuring critically damped behavior along the selected directions.

\subsubsection{Redundancy Resolution}

For the 7-DOF manipulator, kinematic redundancy is exploited in the null space to improve force manipulability along the surface normal. The redundancy torque is defined as
\begin{align}
\bs{\tau}_r &= \bs{N}_M^\top k_{\mathrm{ns}} \nabla_{\bs{q}} \Phi(\bs{q}),\\
\Phi(\bs{q}) &=
\bs{n}^\top
\left(\bs{J}_t\bs{J}_t^\top+\varepsilon\bs{I}_3\right)^{-1}
\bs{n},
\end{align}
where $\bs{N}_M$ is the dynamically consistent null-space projector, $\bs{J}_t$ is the translational Jacobian, and $\bs{n}$ is the surface normal.

The overall commanded torque is
\begin{equation}
\bs{\tau} = \bs{\tau}_f + \bs{\tau}_i + \bs{\tau}_r + \bs{\tau}_g(\bs{q}),
\end{equation}
where $\bs{\tau}_g(\bs{q})$ compensates for gravity.

\section{Force--Impedance Control, GPR, and Ergodic Control parameters}\label{app:control_parameters}
This appendix reports all controller, estimation, and mapping
hyperparameters used in the physical experiments to ensure
full reproducibility.
Both phantoms employ the same Cartesian end–effector impedance
\[
\mathbf{K}_{ee} = [800,\;800,\;0,\;60,\;60,\;30],
\]
while the normal force command differs as summarized in \autoref{tab:force_params}.
\begin{table}[t]
\centering
\caption{Force–impedance and estimator parameters.}
\label{tab:force_params}
\begin{tabular}{lcc}
\toprule
\textbf{Parameter} & \textbf{Hemisphere} & \textbf{Gaus. Dimple} \\
\midrule
Normal force bias [N] & $3$ & $4.25$ \\
Force amplitude [N] & $1.75$ & $2.75$ \\
Excitation frequency & $2$ rad/s & $2$ rad/s \\
PI gains & $[1,\;3]$ & $[1,\;3]$ \\
$\zeta_{\max}$ & $1.003$ & $1.003$ \\
$\theta$ & $1.9$ & $1.9$ \\
Poisson Ratio & $0.5$ & $0.5$ \\
\bottomrule
\end{tabular}
\end{table}
The intrinsic GP used for viscoelastic
mapping adopts the hyperparameters listed in \autoref{tab:gp_params}. The remaining UKF settings are identical to those reported
in~\citet{Beber2025Force-BasedArm}.
\begin{table}[t]
\centering
\small
\caption{GP and ergodic controller hyperparameters, and point cloud initialization.}
\label{tab:gp_params}
\begin{tabularx}{\columnwidth}{@{}Xcc@{}}
\toprule
\textbf{Parameter} & \textbf{Hemisphere} & \textbf{G.\ dimple} \\
\midrule
\multicolumn{3}{@{}l}{\textit{Gaussian process}} \\
Lengthscale $l$ (indenter radius) & \multicolumn{2}{c}{\SI{5}{\milli\meter}} \\
Signal variance $\sigma_f^2$ & \multicolumn{2}{c}{1} \\
Prior mean $m$ & \multicolumn{2}{c}{$1.1\,k_{\min}(t)$} \\
\midrule
\multicolumn{3}{@{}l}{\textit{Point cloud}} \\
Voxel size & \SI{2.0}{\milli\meter} & \SI{1.5}{\milli\meter} \\
\midrule
\midrule
\multicolumn{3}{@{}l}{\textit{Ergodic controller}} \\
Gradient operator neighbors $k$ & \multicolumn{2}{c}{20} \\
Gradient averaging neighbors & \multicolumn{2}{c}{10} \\
Laplacian neighbors & \multicolumn{2}{c}{30} \\
$\alpha$ & \multicolumn{2}{c}{0.4} \\
Diffusion scale $\chi$ & \multicolumn{2}{c}{1} \\
Max.\ velocity $v_{\max}$ & \SI{6}{\milli\meter\per\second} & \SI{5}{\milli\meter\per\second} \\
Max.\ acceleration $a_{\max}$ & \SI{9}{\milli\meter\per\second\squared} & \SI{7}{\milli\meter\per\second\squared} \\
\bottomrule
\end{tabularx}
\end{table}
The prior mean is updated at runtime from the acquired estimates, as
described in~\autoref{sec:surface_gp}.

\section{Silicone Phantom Fabrication}
\label{sec:phantom}
Two custom silicone phantoms with distinct geometries, boundary conditions, and viscoelastic properties were fabricated to evaluate the proposed algorithm.
The selected phantom viscoelasticity range lies within reported values for healthy soft tissues and pathological stiff lesions in palpation-based studies~\citep{wells2011medical}, similarly to related research~\citep{Nichols2015MethodsPalpation,Yan2021FastPalpation,Ayvali2017Utility-GuidedAbnormalities}.
\paragraph{Hemisphere phantom} The first phantom was designed to replicate the geometry and physical characteristics of the commercially available Tokyo Kogaku Ultrasound-Guided Breast Biopsy Phantom (US-9), which is a phantom with realistic acoustic and physical characteristics used for ultrasound-guided breast biopsy training. However, the phantom does not exhibit stiffer lesions, making it unusable for our purposes.
Its bulk matrix was prepared by mixing ECOFLEX-0030 silicone elastomer with silicone oil at a 70\%–30\% weight ratio to reduce stiffness and better approximate the mechanical properties of soft tissue.
Inclusions were manufactured using DRAGONSKIN-30NV, a stiffer silicone formulation.
Three spherical inclusions were embedded within the phantom: two with a radius of \SI{5}{mm} and one with a radius of \SI{1}{cm}.
The resulting phantom exhibits a baseline elasticity of approximately \SI{35}{kPa} for the soft silicone matrix, while the stiffer inclusions present elastic moduli ranging from \SI{50}{kPa} to \SI{80}{kPa}.
The phantom was fabricated through multiple casting stages. After pouring and curing the first silicone layer, two DRAGONSKIN inclusions were positioned inside the mold and encapsulated by an additional ECOFLEX-0030 layer. Once this second layer polymerized, the third inclusion was inserted and embedded by pouring a further matrix layer. Finally, the phantom base was cast using DRAGONSKIN-10 to provide a stiffer supporting substrate, thereby improving handling stability and preventing excessive deformation during robotic palpation experiments.
\autoref{fig:phantom_fabrication} illustrates the resulting phantom and inclusions: \autoref{fig:phantom_fabrication}(a)
shows the first hemispherical phantom, while \autoref{fig:phantom_fabrication}(c) presents the fabricated DRAGONSKIN inclusions used during the manufacturing process, which were colored before insertion in the phantom.

\paragraph{Gaussian Dimple phantom} The Gaussian-dimple phantom provides a challenging complementary case,
as its pronounced concavity and internal boundary test the ability of
the method to preserve intrinsic surface connectivity and maintain
surface-constrained exploration on a non-convex domain. In this case, the bulk matrix consisted solely of ECOFLEX-0030 silicone elastomer.
The phantom incorporated a flat, stiff inclusion of size \SI{1.5}{}x\SI{1.5}{cm}, a spherical inclusion with a radius of \SI{5}{mm}, and a small U-shaped lesion (\SI{2}{}x\SI{5}{mm}), all manufactured using DRAGONSKIN-30NV.
This phantom exhibits a baseline elasticity of approximately \SI{125}{kPa} for the soft silicone matrix, while the stiffer inclusions present elastic moduli ranging from \SI{140}{kPa} to \SI{190}{kPa}.
The resulting Gaussian dimple is shown in \autoref{fig:phantom_fabrication}(b). All inclusions are created using 3D-printed molds.

\end{document}